%% file: main.tex
\documentclass[10pt,twocolumn,letterpaper]{article}
\usepackage[pagenumbers]{cvpr}
\usepackage{iftex}
\ifXeTeX
  \usepackage[OT1]{fontenc}
\fi
\input{preamble}
\usepackage{placeins}
\definecolor{cvprblue}{rgb}{0.21,0.49,0.74}
\usepackage[pagebackref,breaklinks,colorlinks,allcolors=cvprblue]{hyperref}
\usepackage[capitalize]{cleveref}
\crefname{proposition}{Prop.}{Props.}
\hypersetup{
  pdftitle={IDSet-3D: Benchmarking 3D Reconstruction against What the Views Determine},
  pdfauthor={Jialang Liu, Tianfang Tong, Yingmei Wei, Yanming Guo}
}

\title{IDSet-3D: Benchmarking 3D Reconstruction\\against What the Views Determine}
\author{%
Jialang Liu\textsuperscript{1}\quad
Tianfang Tong\textsuperscript{2}\quad
Yingmei Wei\textsuperscript{1}\quad
Yanming Guo\textsuperscript{1,*}\\[4pt]
{\small\textsuperscript{1}National Key Laboratory for Big Data and Decision, National University of Defense Technology (NUDT), Hunan, China}\\
{\small\textsuperscript{2}Changsha University of Science and Technology (CSUST), Hunan, China}\\[3pt]
{\small\href{mailto:liu_1999@nudt.edu.cn}{\texttt{liu\_1999@nudt.edu.cn}}\quad\href{mailto:202403060216@csust.edu.cn}{\texttt{202403060216@csust.edu.cn}}}\\
{\small\href{mailto:weiyingmei@nudt.edu.cn}{\texttt{weiyingmei@nudt.edu.cn}}\quad\href{mailto:guoyanming@nudt.edu.cn}{\texttt{guoyanming@nudt.edu.cn}}
\quad\textsuperscript{*}Corresponding author.}
}

\begin{document}
\twocolumn[{%
\renewcommand\twocolumn[1][]{#1}%
\maketitle
\input{fig/teaser}
}]
\input{sec/0_abstract}
\input{sec/1_intro}
\input{sec/2_related}
\input{sec/3_dataset}
\input{sec/4_benchmark}
\input{sec/5_method}
\input{sec/6_conclusion}

\section*{Data and Model Availability}
The \href{https://huggingface.co/datasets/wolfkingttf/IDSet3D/tree/1ff8c0d1985e31cbaa6e8863ba3ff8003bd61a73}{IDSet-3D data and evaluation outputs} and
\href{https://huggingface.co/wolfkingttf/IDSet3D-IDHead/tree/570a01bd03eb289f2fe2a9778ad5f19af10f4ed2}{ID-Head checkpoints} are available on Hugging Face. The links point to the revisions used for this preprint. See \cref{supp:data} for the released files.

\section*{LLM Usage Statement}
We used AI code writing tools like Codex and Claude Code, which also helped us run experiments, organize data, generate figures and tables, and discuss how to implement parts of the method. We also leveraged ChatGPT and Claude to search related work and to refine phrasing in some areas for improved clarity. All work and ideas are our own.

{
    \small
    \bibliographystyle{ieeenat_fullname}
    \bibliography{main}
}

\clearpage
\appendix
\crefname{appendix}{Appendix}{Appendices}
\Crefname{appendix}{Appendix}{Appendices}
\setcounter{table}{0}
\setcounter{figure}{0}
\renewcommand{\thetable}{S\arabic{table}}
\renewcommand{\thefigure}{S\arabic{figure}}
\renewcommand{\theHtable}{supp.\arabic{table}}
\renewcommand{\theHfigure}{supp.\arabic{figure}}
\section*{Supplementary Material}
\noindent \Cref{supp:prereg} lists every registered hypothesis with its outcome and documents every deviation from the registered protocol. \Cref{supp:eval} specifies the evaluation pipeline and illustrates how the evidence and the true first hit of each ray give its endpoint state. \Cref{supp:bench} presents the complete test-1 benchmark table, constructed reference answers, counterfactual twins, results by query stratum, depth and per-ray judgment examples, the test-2 replication, window-level statistics, and sensitivity to the evaluation settings. \Cref{supp:head} documents \head and its development---including a validation gate it did not pass---along with where it abstains and fails and its secondary results. \Cref{supp:data} describes the dataset splits, shows one window from each split, and outlines the public release.

\input{supp/A_registration}
\input{supp/B_evaluation}
\input{supp/C_benchmark}

\input{supp/D_head}

\input{supp/E_data}
\FloatBarrier

\end{document}

%% file: preamble.tex
\usepackage{cuted} 

\usepackage{currfile} 

\usepackage{caption} 

\usepackage{booktabs}
\usepackage{xcolor}
\usepackage{amsthm}
\usepackage{multirow}

\providecommand{\emptycell}{\textemdash}

\usepackage{pifont}
\newcommand{\cmark}{\ding{51}}
\newcommand{\xmark}{\ding{55}}

\usepackage{array}
\usepackage{colortbl}
\usepackage{makecell}
\usepackage{adjustbox}
\definecolor{tabours}{HTML}{D9E7F4}     
\definecolor{tabgroup}{HTML}{F2F2F2}    
\definecolor{tabbest}{HTML}{C00000}     
\definecolor{tabmet}{HTML}{BCE2C3}      
\definecolor{tabnotmet}{HTML}{FFC7CE}   
\definecolor{tabopen}{HTML}{FFF9D9}     
\definecolor{tabnote}{HTML}{6E6E6E}     
\newlength{\tabframewidth}
\newcolumntype{?}{!{\vrule width \tabframewidth}}  
\newcommand{\tabsetup}{\footnotesize\setlength{\tabcolsep}{3pt}\renewcommand{\arraystretch}{1.1}}
\newcommand{\tabtop}{\Xhline{\tabframewidth}}      
\newcommand{\tabhead}{\Xhline{\tabframewidth}}     
\newcommand{\tabbottom}{\Xhline{\tabframewidth}}   
\newcommand{\tabgap}{\noalign{\vskip 3pt}}         
\newcommand{\bestval}[1]{\textcolor{tabbest}{#1}}
\newcommand{\secondval}[1]{\underline{#1}}
\newcommand{\venue}[1]{\,{\scriptsize\color{tabnote}[#1]}}
\newcommand{\ci}[1]{{\scriptsize #1}}                
\newcommand{\cellmet}{\cellcolor{tabmet}}
\newcommand{\cellnotmet}{\cellcolor{tabnotmet}}
\newcommand{\cellopen}{\cellcolor{tabopen}}
\newcolumntype{C}[1]{w{c}{#1}}                       
\newcommand{\tabfirst}[2]{\multicolumn{1}{?c|}{\multirow{#1}{*}{#2}}}  
\newlength{\tabwidth}                                
\newcommand{\tabnotes}[1]{\\[2pt]\parbox{\tabwidth}{\scriptsize #1}}  

\newtheorem{proposition}{Proposition}

\newcommand{\inputs}{\mathcal{O}}          
\newcommand{\eqclass}{\mathcal{E}}         
\newcommand{\surf}{\mathcal{S}}            
\newcommand{\free}{\mathcal{F}}            
\newcommand{\idset}{\Omega}                
\newcommand{\dstar}{d^{*}}                 
\newcommand{\dhat}{\hat{d}}                
\newcommand{\tauv}{\tau}                   
\newcommand{\stV}{\textsf{V}\xspace}       
\newcommand{\stH}{\textsf{H}\xspace}       
\newcommand{\stN}{\textsf{N}\xspace}       
\newcommand{\stX}{\textsf{X}\xspace}       
\newcommand{\stE}{\textsf{E}\xspace}       
\newcommand{\pithree}{$\pi^3$\xspace}
\newcommand{\bench}{\mbox{IDSet-3D}\xspace}   
\newcommand{\head}{\mbox{ID-Head}\xspace}

%% file: fig/teaser.tex
\begingroup
\centering
\captionsetup{type=figure}
\includegraphics[width=0.88\textwidth]{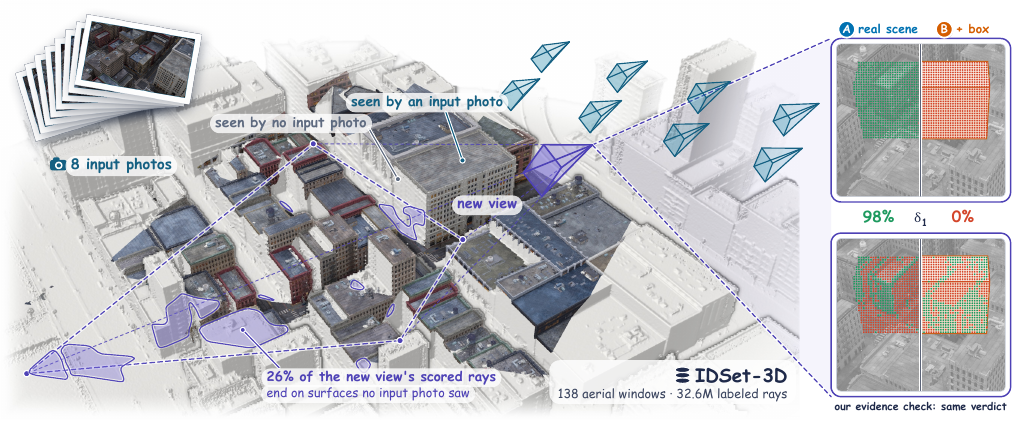}
\captionof{figure}{\textbf{Scoring against what the inputs determine.} The ground-truth city of one \bench window keeps its color where one of its eight input photos (teal) saw it, and 26\% of the scored rays of a held-out new view end on surfaces that no input photo saw (violet). World~B adds a box that the new view sees and no input photo does, so every model predicts one depth for both worlds. On the 1{,}743 rays it hides (one dot each), $\delta_1$ of the \mbox{VGGT-1B} prediction is 98\% in world~A and 0\% in world~B; our evidence check gives the same verdict in both (red: fails $\delta_1$ or contradicted). Over 40 certified twins, the median $\delta_1$ of the seven answering models falls from 0.22--0.33 to 0.00.}
\label{fig:teaser}
\par\endgroup

%% file: sec/0_abstract.tex
\begin{abstract}
In sparse aerial reconstruction, novel viewpoints reveal surfaces that none of the input cameras observed. Their depths remain ambiguous, but predictions can still contradict the input evidence. Standard depth thresholds can accept predictions that place a surface in space an input camera sees through, and visibility masks can omit the same errors when the true surface is hidden. We propose \bench{}, a benchmark that evaluates novel-view depth against a set-valued reference derived from exact, posed input-view depth maps. The benchmark comprises 138 synthetic MatrixCity windows with eight input images each. For each query ray, the reference specifies which first-hit depths the input evidence rules out, exposing contradictions even where the true endpoint is unobserved. On the primary test block, six of seven counted models place more than 10\% of their answers on visibility-masked rays in certified free space at the median query, and 81--92\% of these violations still pass the standard depth threshold ($\delta_1$). Across 40 counterfactual scene pairs with equivalent input depth observations, median $\delta_1$ on rays whose reference depths change falls from 0.22--0.33 in the original scenes to 0.00 for all seven models in the modified scenes. The same predictions receive identical evidence verdicts. The endpoint labels also support selective reconstruction. On a frozen VGGT-1B, \head{}, a 0.35M-parameter selector, reduces evidence-violating hidden-endpoint answers by 22--37\% relative to the strongest training-free gate across two test blocks, averaged over matched observed-endpoint recalls from 0.30 to 0.90.
\end{abstract}
\vspace{-5.5pt}

%% file: sec/1_intro.tex
\section{Introduction}
\label{sec:intro}

A novel-view depth prediction can pass a standard accuracy threshold while placing a surface in space that an input camera has already seen through. If the true endpoint was never observed, a visibility mask can discard the ray altogether. Occlusion makes the target ambiguous; it does not make every prediction defensible. Modern 3D models reconstruct scenes from a handful of images in one forward pass~\cite{wang2024dust3r,leroy2024mast3r,wang2025vggt,wang2025pi3,keetha2025mapanything,lin2025da3}. Their geometry is evaluated against held-out ground-truth depth using AbsRel and $\delta_1$~\cite{eigen2014depth}. These scores measure agreement with the realized scene; they do not separately identify contradictions of the input evidence.

A counterfactual twin (\cref{fig:teaser}) makes this distinction concrete. World~B adds a box unseen by the input cameras to world~A. Input depths agree; images are reused assuming no visible shadow or reflection, so each model makes the same prediction. On the affected rays, \mbox{VGGT-1B}~\cite{wang2025vggt} earns $\delta_1 = 0.98$ against world~A and $0.00$ against world~B. Across 40 certified twins, median $\delta_1$ is 0.22--0.33 in world~A and falls to 0.00 in world~B for all seven answering models. Scene accuracy changes with the reference world while input evidence and predictions remain fixed.

Current protocols either score unobserved surfaces against one completion~\cite{li2025amodal,akola2026z3d} or omit rays whose true endpoint was not observed~\cite{warburg2023nerfbusters,tian2023occ3d} (\cref{tab:protocols}). Neither choice separates ambiguity about the endpoint from constraints on the prediction. A hidden endpoint can lie behind space that an input camera sees through. Placing a surface in that space contradicts the evidence, regardless of the hidden depth. Evaluating such predictions requires retaining the geometric constraints along each ray, even when its endpoint is unobserved.

We introduce \bench, a benchmark with a set-valued reference derived from exact, posed input-view depth maps. This \emph{identified set} follows partial identification in econometrics~\cite{manski2003partial} and the photo hull of shape from images~\cite{kutulakos2000spacecarving}: it retains candidate depths that the input geometry cannot rule out. Along each query ray, input depths certify observed surfaces and free space. A prediction violates the evidence if it lies in certified free space or behind the first observed surface; consistency alone does not establish correctness. For exact evidence, these per-ray judgments remain unchanged across worlds with equivalent input-depth observations (\cref{prop:invariance}), a property checked by the twins. Models receive images and, where required, cameras; only the evaluator reads reference depths. Query ground truth separately labels \emph{observed-endpoint} and \emph{hidden-endpoint} rays, distinguishing agreement with observed and unobserved surfaces. Built from MatrixCity aerial renderings~\cite{li2023matrixcity}, \bench provides exact geometry for computing these references and spatially held-out blocks for evaluation (\cref{fig:dataset}).

We evaluate eight models spanning feed-forward multi-view reconstruction, monocular metric depth, and 3D Gaussian splatting~\cite{kerbl20233dgs}. On the primary test block, six of seven counted models place over 10\% of a median query's answers on visibility-masked rays in certified free space; 81--92\% of these violations still pass $\delta_1$. Hidden endpoints also account for 14--24\% of standard $\delta_1$ passes, contributing 0.08--0.15 to the score, more than the entire spread across models. Both statistics were registered after exploring four models, confirmed on three unseen models, and replicated on a second block. Evidence consistency exposes errors that accuracy thresholds accept and visibility masks omit.

The endpoint labels support learning when to answer. \head, a 0.35M-parameter U-Net using 18 features of frozen \mbox{VGGT-1B} predictions (\cref{fig:overview}), learns to select observed endpoints. At matched observed-endpoint recall, it returns 22--25\% (\mbox{test-1}) and 29--37\% (\mbox{test-2}) fewer evidence-violating hidden-endpoint answers than the strongest of six training-free gates, including model confidence, averaged over the registered recall range. The improvement holds across all five seeds on both test sets.

\noindent\textbf{Contributions.}
(i)~\bench, a set-valued evaluation benchmark with invariant per-ray evidence judgments, certified counterfactual twins, and 138 labeled MatrixCity windows containing 32.6M rays and spatially held-out test blocks (\cref{sec:dataset}).
(ii)~A study of eight models with registered confirmation and replication, exposing evidence violations missed by depth thresholds and visibility masks, and quantifying accuracy credit on hidden endpoints (\cref{sec:benchmark}).
(iii)~\head, demonstrating that observed-endpoint training labels support selective reconstruction with fewer evidence-violating answers at matched observed-endpoint recall than every compared training-free gate on both test sets (\cref{sec:method}).

%% file: sec/2_related.tex
\section{Related Work}
\label{sec:related}

\paragraph{Scoring reconstructed geometry.}
Feed-forward reconstruction models are scored by depth and point-map error against a single ground-truth world~\cite{wang2024dust3r,leroy2024mast3r,wang2025vggt,wang2025pi3,keetha2025mapanything,lin2025da3}, with monocular depth metrics~\cite{eigen2014depth,bochkovskii2025depthpro,wang2025moge2}, although glass makes the depth label a convention~\cite{xu2026onescene}. Rendered cities supply such worlds at aerial scale~\cite{li2023matrixcity}, and Sky2Ground benchmarks the poses of DUSt3R, MASt3R, MapAnything, and VGGT across ground, aerial, and satellite views~\cite{wang2026sky2ground}. Unobserved space is excluded by a visibility mask~\cite{warburg2023nerfbusters,tian2023occ3d,tseng2026aligning} or scored against one completion~\cite{li2025amodal,akola2026z3d}, and ETH3D penalizes points in space its ground-truth scanner saw through~\cite{schoeps2017eth3d}. All of these keep a single world, or a sensor that saw more than the inputs, as the reference (\cref{tab:protocols}). \bench penalizes only answers the inputs rule out, so each penalty holds in every world they allow.
\vspace{-5.5pt}

\input{generated/tab1_protocols}

\paragraph{Uncertainty and identifiability.}
Uncertainty estimates for radiance fields~\cite{goli2024bayesrays,shen2022cfnerf,jiang2024fisherrf,gavis2026,gottwald2026primu} describe a model and are evaluated against one world. The identified set is fixed by the inputs before any model runs. That images determine a scene only up to a set is classical: the visual hull and the photo hull bound the shapes consistent with silhouettes and with photographs~\cite{laurentini1994visualhull,kutulakos2000spacecarving}, and econometrics treats such ambiguity as partial identification~\cite{manski2003partial}. Separating what the data determine from estimation error also shapes evaluation in other domains, from embodied question answering that penalizes answers whose evidence is not visible~\cite{e3vs2026} to null-space hallucinations in tomography~\cite{bhadra2021hallucinations}. Plausible completions~\cite{wu2020multimodal} and multi-view consistency~\cite{asim2025met3r} have been studied without a set-valued reference.
\vspace{-5.5pt}

\paragraph{Selective prediction.}
The reject option~\cite{chow1970optimum} has been developed for deep networks through confidence thresholds and learned reject heads~\cite{geifman2017selective,geifman2019selectivenet}. In 3D, reconstruction confidence routinely filters points~\cite{wang2024dust3r,wang2025vggt}, and learned readouts on frozen feed-forward backbones predict pose failure~\cite{srinivasan2026vggtknow,ahmedtaristizabal2026geocond}, per-point uncertainty~\cite{zhu2026trust3r}, or co-visibility~\cite{ziliotto2026covggt}. These readouts target failure, uncertainty, or overlap. \head learns to select observed endpoints: a hidden-endpoint ray remains a negative training example even when its predicted depth matches the query ground truth.

%% file: generated/tab1_protocols.tex
\begin{table}[t]
\centering
\caption{\textbf{Comparison with evaluation protocols.} How each protocol treats a ray whose endpoint no input observed, which observations make its flagged errors certain, and whether its verdict is the same in every world consistent with the inputs.}
\label{tab:protocols}
\tabsetup
\begin{adjustbox}{max width=\columnwidth}
\begin{tabular}{?l|c|c|c?}
\tabtop
Protocol & \makecell{Unobserved\\endpoint} & \makecell{Errors certain\\given} & \makecell{Same\\verdict} \\
\tabhead
Depth metrics~\cite{eigen2014depth,wang2025vggt} & one world & -- & \xmark \\
Visibility masks~\cite{warburg2023nerfbusters,tian2023occ3d,tseng2026aligning} & not scored & -- & \xmark \\
Completion targets~\cite{li2025amodal,akola2026z3d} & one completion & -- & \xmark \\
Scanner free space~\cite{schoeps2017eth3d} & not scored & the scanner & \xmark \\
\rowcolor{tabours} \bench (ours) & identified set & the inputs & \cmark \\
\tabbottom
\end{tabular}
\end{adjustbox}
\end{table}

%% file: sec/3_dataset.tex
\section{The \texorpdfstring{\bench}{IDSet-3D} Dataset}
\label{sec:dataset}

Each sample of \bench is a \emph{window}: eight input views with images, cameras, and depth, paired with up to 24 held-out query views of the same area (\cref{fig:dataset}b). Every query ray carries the evidence that the inputs certify along it and an endpoint state derived from the query ground truth (\cref{fig:dataset}c).

\input{fig/dataset}

\subsection{Scenes, pairing, and splits}
\label{sec:pairing}

\paragraph{Source scenes.}
MatrixCity~\cite{li2023matrixcity} renders a synthetic city with exact depth and poses. Its aerial captures are $1920{\times}1080$ frames at $45^\circ$ pitch and $45^\circ$ field of view on a grid of camera positions, each with four yaws (\cref{fig:dataset}b). After converting positions and depth to meters, neighboring frames reproject onto each other with an error of 0.014\%.
\vspace{-5.5pt}

\paragraph{Input--query pairing.}
A window pairs $K = 8$ input views, at one yaw from a $2{\times}4$ block of grid positions, with query views at the same yaw within two grid steps. Query views are stratified without query ground truth, by the fraction $\rho_b$ of their rays with a finite upper bound $b$ (\cref{sec:evidence}): interpolation ($\rho_b \ge 60\%$), extrapolation (30--60\%), or low evidence (10--30\%), at most eight views each. Each query view is evaluated on a $90{\times}160$ grid of native pixels (14{,}400 rays) with a range of 500\,m at 150\,m altitude.

\paragraph{Splits.}
As in spatially partitioned cross-view datasets~\cite{xu2025sat2groundscape}, the test windows come from city blocks that no training or validation window uses (\cref{fig:dataset}a, with the size and label composition of each split in \cref{fig:dataset}d and \cref{supp:data}). Training and validation windows (blocks 9 and 10, 200--250\,m) supply labels for \head. \mbox{Test-1} (block~1) is the primary benchmark, and \mbox{test-2} (block~2) was registered later as a replication. No frame is shared between splits, and validation windows lie at least six grid steps from training windows. Two further block-1 windows, frame-disjoint from \mbox{test-1}, served for development (the inclusion check, the evidence-wrapper parameters, and the method protocol). Block~2 entered no development step.

\subsection{Identified-set annotation}
\label{sec:evidence}

\input{fig/overview}

\paragraph{Observations and equivalent worlds.}
A world $w$ is a collection of opaque surfaces. The input $\inputs$ consists of $K$ views with known pinhole cameras. View $k$ records an image $I_k$ and, at each pixel $p$, the depth $Z_k(p)$ of the first surface within range on the ray through the pixel center, or $+\infty$. Models receive the images (and the cameras where they require them), and only the evaluator reads $Z_k$. Two worlds are \emph{observationally equivalent} if they produce the same $Z_k(p)$ at every input pixel, and $\eqclass(\inputs)$ is the class of worlds equivalent to the rendered one. A query ray is $r(t) = o + t\,v$, $t \in [0, T]$, with first hit $\dstar_w(r)$ in world $w$. Because equivalence is defined through depth, the identified set captures the ambiguity that remains even when every input view is reconstructed perfectly.
\vspace{-5.5pt}

\paragraph{Evidence along a query ray.}
The inputs certify two sets (\cref{fig:overview}, top). A point $x$ belongs to the \emph{observed surface} $\surf$ if some input view $k$ sees it at the recorded depth, and to the \emph{certified free space} $\free$ if some input view sees through it to a farther surface:
\begin{equation}
\bigl|z_k(x) - Z_k(p)\bigr| \le \tau_s(z_k(x)), \qquad z_k(x) < Z^{\min}_k(p) - \tau_s(z_k(x)),
\label{eq:surface}
\end{equation}
where $p = p_k(x)$ and $z_k(x) > 0$ are the projection and camera depth of $x$ in view $k$, $\tau_s(z) = \max(0.05\,\mathrm{m}, 0.01\,z)$, and $Z^{\min}_k$ is the input depth after block-minimum downsampling and a $3{\times}3$ minimum filter, which keeps $\free$ conservative at depth edges. The \emph{upper bound} $b(r)$ is the distance to the first point of $\surf$ along $r$ within range (computed on a depth mesh of each input view; \cref{supp:eval}), or $+\infty$. With $L = \min(b, T)$, the \emph{identified set} of first-hit distances is
\begin{equation}
\begin{split}
\idset(r) = {}&\{ t \in [0, L] : r(t) \notin \operatorname{int}\free \}\\
&\cup \{+\infty \text{ if } b = +\infty\}.
\end{split}
\label{eq:idset}
\end{equation}
We sample $[0, L]$ at $N = 256$ midpoints with spacing $\Delta$ and use the tolerance $\tauv(d) = 0.05\,\mathrm{m} + 0.01\,d + 2\Delta$. The length of $[0,L]$ outside $\free$ is the uncertain length $u$, and $u/L$ is the evidence gap.

\paragraph{Judging a prediction.}
A prediction $\dhat \in (0, +\infty]$ contradicts the evidence in two ways:
\begin{equation}
\begin{aligned}
\text{front} &\colon\quad \dhat < \infty \ \text{and}\ r(\dhat) \in \free,\\
\text{back} &\colon\quad b < \infty \ \text{and}\ \dhat > b + \tauv(b).
\end{aligned}
\label{eq:violations}
\end{equation}
A front violation uses the free-space margin $\tau_s(z)$ in \cref{eq:surface} and places a surface where an input view sees through. It is the free-space violation of visibility-based depth-map fusion~\cite{merrell2007fusion}, used here to judge a prediction rather than to fuse depth. A back violation places a surface behind an observed one, or declares the ray empty although it reaches one. Every other valid answer is \emph{consistent} under the evaluator's numerical rules. In the uncertain part of $\idset(r)$, the inputs exclude no answer and confirm none.

\begin{proposition}[Soundness and invariance]
\label{prop:invariance}
For exact evidence computed from continuous input depth maps without tolerances or discretization:
(a)~every $w \in \eqclass(\inputs)$ satisfies $\dstar_w(r) \in \idset(r)$ for every query ray $r$;
(b)~for any fixed prediction, the indicators in \cref{eq:violations} are the same for every $w \in \eqclass(\inputs)$.
\end{proposition}

\begin{proof}
(a)~If $b < \infty$, $r(b)$ is the first hit of an input pixel ray; every $w \in \eqclass(\inputs)$ reproduces that depth, so $\dstar_w(r) \le b$. A surface of $w$ in $\free$ would shorten the depth of the input pixel that sees through it, so $r(\dstar_w) \notin \operatorname{int}\free$.
(b)~The indicators depend only on $\surf$, $\free$, $b$, $\tauv$, and the prediction, which are functions of the inputs.
\end{proof}

Under the exact assumptions in (a), either violation rules out the predicted first-hit depth in every world consistent with the input depth observations. Single-world metrics lack property (b): AbsRel and $\delta_1$ use $\dstar_w$, which varies across $\eqclass(\inputs)$.

\subsection{Endpoint states}
\label{sec:states}

Endpoint states describe the rendered first hit $\dstar$ relative to the input-depth evidence. They depend on the rendered scene and the input views. For a finite hit, \stV (observed endpoint) indicates a finite bound $b$ with $|\dstar-b|\leq\tauv(b)$. The label \stH (hidden endpoint) indicates either $b=+\infty$ or $\dstar<b-\tauv(b)$ for finite $b$. We use \stN for rays with no rendered hit. A finite hit is flagged as \stX (evidence-inconsistent) if it exceeds a finite bound by more than $\tauv(b)$ or lies in $\operatorname{int}\free$. A query pixel is flagged as \stE (edge-ambiguous) if it lies within 2 pixels of a depth jump of at least 2\%, where rendered depth can mix foreground and background. We exclude \stE and \stX rays from all metrics. Matching $\dstar$ on a \stV ray recovers an observed surface. Matching $\dstar$ on an \stH ray agrees with the rendered scene at an unobserved surface, where the input evidence can still rule out other answers.
\vspace{-5.5pt}

\paragraph{Ambiguity of the identified set.}
The identified set contains multiple candidate depths on every non-excluded test ray. Each such ray has an estimated evidence gap $u/L$ above 0.10, and about half have no finite upper bound. In this aerial setup, query cameras lie outside the input frusta, leaving space near the ray origin uncertified even when the ray eventually reaches an observed surface. Scoring only rays with a singleton reference set would therefore leave no test rays to score. Evidence violations remain measurable on these rays.

\subsection{Testing invariance with counterfactual twins}
\label{sec:twins}

\Cref{prop:invariance} establishes invariance for exact evidence. Counterfactual twins test whether the implemented per-ray verdict stays fixed when the query reference depth changes and the input evidence and prediction are held fixed. World~B is constructed by adding an axis-aligned box with side lengths of 3--15\,m to a test window from world~A. The box is centered on a query ray in the uncertain part of $\idset$, before the original first hit. A twin is accepted if (i)~no input pixel ray enters the box more than $\tauv$ before its recorded depth, (ii)~all 2{,}000 points sampled inside the box lie outside both $\surf$ and $\free$, (iii)~the box contains no surface of~A, (iv)~the box contains no camera, and (v)~at least one query ray enters the box more than $\tauv$ before its first hit in~A. The box stays at least 0.5\,m from $\surf$. World~B reuses the input images from~A under the assumption that the inserted box casts no visible shadow or reflection. The query reference in~B is $\dstar_B=\min(\dstar_A,t_{\mathrm{box}})$, where $t_{\mathrm{box}}$ is the ray's first intersection with the box, or $+\infty$ if it misses. The \emph{divergence region} $D$ consists of query rays with $|\dstar_A-\dstar_B|>\tauv$. The same model prediction is scored against both query references on $D$ and checked against the shared input-depth evidence.

%% file: fig/dataset.tex
\begin{figure*}[t]
\centering
\includegraphics[width=\textwidth]{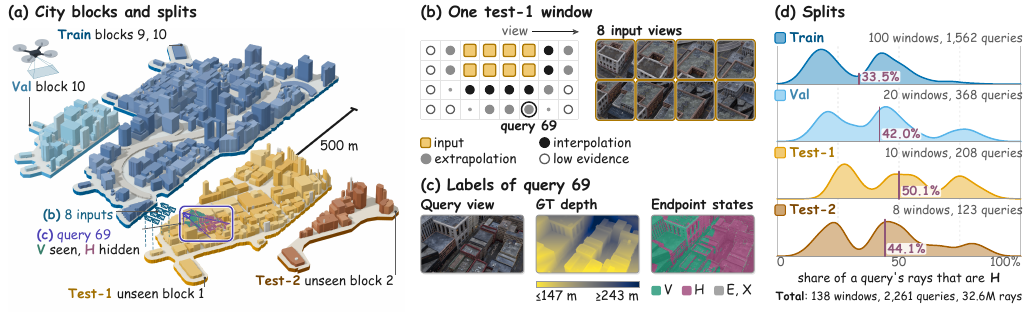}
\caption{\textbf{Overview of the \bench dataset.} (a)~The city under the four aerial blocks we use (heights from query-view depth), cut into the areas of the splits and drawn apart (colors as in (d)), with the cameras of (b) and the ray endpoints of the query of (c). (b)~A \mbox{test-1} window: eight input views at one yaw on a $2{\times}4$ block of camera positions, and the query views within two grid steps (38\,m each), by evidence stratum. (c)~One query with its ground-truth depth and endpoint states. (d)~Per split, the share of H rays in each query (mark: over all rays).}
\label{fig:dataset}
\end{figure*}

%% file: fig/overview.tex
\begin{figure*}[t]
\centering
\includegraphics[width=\textwidth]{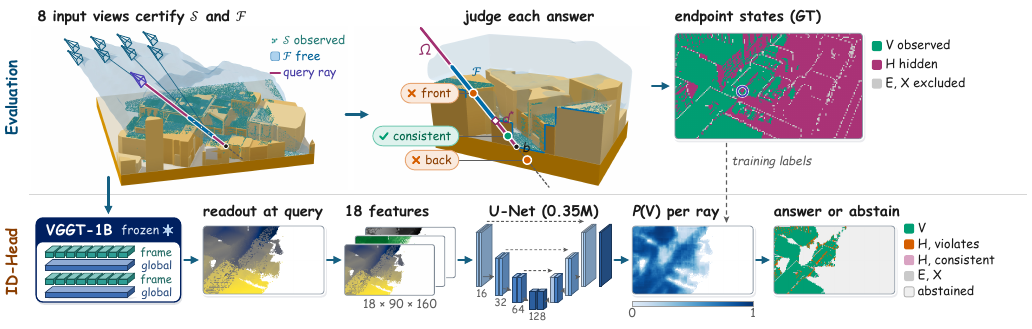}
\caption{\textbf{Overview of \bench and \head.} \emph{Evaluation} (\cref{sec:evidence,sec:states}): input views certify the surface they observe ($\mathcal{S}$) and the space they see through ($\mathcal{F}$), on the model of \cref{fig:dataset}a, cut along a query ray. On it, $\Omega$ holds the depths up to the bound $b$ outside $\mathcal{F}$; an answer in $\mathcal{F}$ is a front violation, one behind $b$ a back violation, any other consistent. Ground truth ($d^*$) only assigns endpoint states. \emph{\head} (\cref{sec:method}): a frozen \mbox{VGGT-1B} reconstructs the scene; its readout and the evidence from its depths give 18 features per ray, and a U-Net predicts $P(\stV)$, thresholded to answer or abstain. Endpoint states supervise training only. Query 0058 of \texttt{mc\_yaw0\_ix3\_iy0}.}
\label{fig:overview}
\end{figure*}

%% file: sec/4_benchmark.tex
\section{The \texorpdfstring{\bench}{IDSet-3D} Benchmark}
\label{sec:benchmark}

\paragraph{Experimental setup.}
\mbox{Test-1} is the primary benchmark and \mbox{test-2} a replication on a block unused in development. The eight models, none trained on \bench, receive the eight input images and no query image. The evaluator, statistics, their criteria, and method protocol in \cref{sec:method} were registered before the corresponding test data were scored, and \cref{supp:prereg} lists every hypothesis and deviation. The 95\% intervals use 2{,}000 window-bootstrap resamples.
\vspace{-5.5pt}

\paragraph{Models.}
We evaluate five feed-forward multi-view models (\mbox{VGGT-1B}~\cite{wang2025vggt}, \pithree~\cite{wang2025pi3}, MapAnything~\cite{keetha2025mapanything}, Depth Anything~3 (DA3)~\cite{lin2025da3}, MASt3R~\cite{leroy2024mast3r}), two monocular metric-depth models (\mbox{MoGe-2}~\cite{wang2025moge2}, Depth Pro~\cite{bochkovskii2025depthpro}), and per-scene 3D Gaussian splatting (3DGS)~\cite{kerbl20233dgs} initialized with COLMAP~\cite{schoenberger2016colmap}. Predicted input point maps are aligned to the input cameras by a similarity transform~\cite{umeyama1991} and splatted into the query camera using a $4{\times}$ supersampled z-buffer. Empty pixels answer $+\infty$, and 3DGS renders depth directly (\cref{supp:eval}). A preregistered inclusion check on the development windows (input-view $\delta_1 \ge 0.80$, coverage $\ge 80\%$) excludes Depth Pro, whose metric depth is about $51{\times}$ too small here. It is reported but not counted.
\vspace{-5.5pt}

\paragraph{Metrics.}
The standard metrics, $\delta_1$ and AbsRel, compare a prediction with the depth of the rendered world on the scored rays (\stE and \stX excluded). Two statistics split this credit by endpoint state and count answers that the inputs rule out. A model's \emph{hidden share} is the fraction of its $\delta_1$ passes that occur on \stH rays (pooled within a window, median over windows). The \emph{free-space answer rate} is the fraction of a query's answered \emph{masked} rays, the rays a visibility mask would skip because their true endpoint is not observed, that are front violations (median over queries, then windows). Both statistics were registered after scoring four exploration models (\mbox{VGGT-1B}, \pithree, MapAnything, \mbox{MoGe-2}) on \mbox{test-1}. They were then tested on three unseen confirmation models (DA3, MASt3R, 3DGS), with the criterion that the statistic reach 10\% (hidden share) or exceed it (free-space answers) for at least half the counted models, and finally on \mbox{test-2} with all seven counted models. They follow four hypotheses registered before any model ran, three of whose definitions do not fit aerial novel views (one not computable, three not met), and a ranking-change and an answer-state hypothesis of the same round were not confirmed.

\input{generated/tab2_findings}

\subsection{What standard depth evaluation misses}
\label{sec:findings}

\paragraph{Visibility masks hide free-space violations.}
On \mbox{test-1}, 14.5--39.6\% of answers on visibility-masked rays fall in certified free space for six of the seven counted models, using the median over queries and then windows (\cref{tab:findings}). These predictions place a surface where an input view sees through, even though the true endpoint is hidden. The median violating prediction lies 5--8\% in front of the hidden surface, and 81--92\% of the violations still pass $\delta_1$. Halving or doubling both tolerance margins keeps all six models above 10\% (\cref{supp:bench}). Two of the three confirmation models exceed the registered threshold (DA3 at 33.1\% and MASt3R at 27.3\%). The criterion is also met on \mbox{test-2}, where four of seven models exceed 10\% (\mbox{MoGe-2}, DA3, MASt3R, and MapAnything) and the other three range from 4.1\% to 6.0\% (\cref{fig:findings}b). The criterion remains met when any single model is omitted.
\vspace{-5.5pt}

\paragraph{3DGS tends to predict beyond hidden surfaces.}
On \mbox{test-1}, 3DGS has a free-space violation rate of 7.3\% on answered masked rays. On hidden-endpoint rays, its median predicted depth is 2.58 times the reference depth. This pattern is consistent with its per-scene optimization against the input views, which constrains geometry along those views.
\vspace{-5.5pt}

\paragraph{Standard $\delta_1$ includes agreement with hidden surfaces.}
Hidden-endpoint rays make up 54\% of scored rays in the median \mbox{test-1} window. Models pass $\delta_1$ on 15--28\% of these rays, accounting for 14--24\% of all passes (\cref{tab:findings}). Only 1.5--4.6\% of these passing predictions lie within 5\% of an observed surface behind the hidden endpoint. Passes on hidden endpoints contribute an absolute 0.08--0.15 to each model's overall $\delta_1$, exceeding the 0.061 spread across the seven counted models. The rankings differ between the two metrics. \pithree ranks first on observed endpoints ($\delta_1 = 0.984$) and sixth overall. MapAnything ranks last on observed endpoints (0.897) and second overall, with the highest pass rate on hidden endpoints (0.277). All three confirmation models meet the registered criterion (19.4--22.8\%). On \mbox{test-2}, six of seven models reach the 10\% threshold, with hidden shares of 10.1--21.6\% among those six (\cref{fig:findings}a).
\vspace{-5.5pt}

\input{fig/findings}

\paragraph{Changing the reference scene changes accuracy scores.}
The 40 counterfactual twins, four per \mbox{test-1} window, isolate the effect of changing the query reference with input evidence and predictions held fixed. The median twin changes the first hit on 1{,}160 query rays, less than 0.5\% of its window. On these rays, median $\delta_1$ ranges from 0.22 to 0.33 in world~A across the seven models that answer and falls to 0.00 for all seven in world~B. For the learned models, AbsRel rises from 0.03--0.11 to 2.0--2.3 (\cref{fig:teaser}, \cref{supp:bench}). The per-ray evidence verdicts agree across both worlds for all eight models and all 40 twins. For any fixed depth prediction, the larger of its two absolute errors is at least half the gap between the reference depths. On rays answered by the learned models, this lower bound is 63--64\,m, and the measured worse-world errors are 123--141\,m.

Together, these results show why depth accuracy and evidence consistency are complementary. Depth accuracy measures agreement with the rendered scene, while evidence consistency exposes contradictions that can otherwise pass the depth threshold or be omitted by a visibility mask. The training windows carry the same annotations, which \cref{sec:method} uses for selective answering.

%% file: generated/tab2_findings.tex
\begin{table}[t]
\centering
\caption{\textbf{Where $\delta_1$ comes from, and answers in certified free space.} $\delta_1$: test-1 window medians over all scored, \stV, and \stH rays. \emph{Hidden share}: fraction of $\delta_1$ passes on \stH rays. \emph{Free-space}: fraction of masked-ray answers that are front violations (median query). Both in \%. Gray: confirmation models. Last row: counted models that meet the registered 10\% criterion. Columns with an arrow: best in red, second underlined.}
\label{tab:findings}
\tabsetup
\setlength{\tabcolsep}{2.8pt}
\begin{adjustbox}{max width=\columnwidth}
\begin{tabular}{?l|ccc|cc|cc?}
\tabtop
\multirow{2}{*}{Model} & \multicolumn{3}{c|}{$\delta_1$ (test-1)} & \multicolumn{2}{c|}{Hidden share} & \multicolumn{2}{c?}{Free-space $\downarrow$} \\
\cline{2-8}
 & all $\uparrow$ & \stV $\uparrow$ & \stH & test-1 & test-2 & test-1 & test-2 \\
\tabhead
VGGT-1B~\cite{wang2025vggt} & 0.542 & \secondval{0.970} & 0.173 & 15.6 & 12.7 & 19.9 & 6.0 \\
\pithree~\cite{wang2025pi3} & 0.534 & \bestval{0.984} & 0.157 & 13.9 & 9.3 & 15.0 & \bestval{4.1} \\
MapAnything~\cite{keetha2025mapanything} & \secondval{0.573} & 0.897 & 0.277 & 24.4 & 21.6 & \secondval{14.5} & 12.7 \\
MoGe-2~\cite{wang2025moge2} & 0.517 & 0.969 & 0.153 & 14.9 & 10.1 & 39.6 & 31.3 \\
\rowcolor{tabgroup} DA3~\cite{lin2025da3} & 0.567 & 0.969 & 0.246 & 22.4 & 17.8 & 33.1 & 26.2 \\
\rowcolor{tabgroup} MASt3R~\cite{leroy2024mast3r} & \bestval{0.578} & 0.939 & 0.240 & 22.8 & 15.4 & 27.3 & 23.4 \\
\rowcolor{tabgroup} 3DGS~\cite{kerbl20233dgs} & 0.537 & 0.918 & 0.214 & 19.4 & 16.9 & \bestval{7.3} & \secondval{5.7} \\
\rowcolor{tabours} \multicolumn{4}{?l|}{Counted models meeting the criterion} & 3/3 & 6/7 & 2/3 & 4/7 \\
\tabbottom
\end{tabular}
\end{adjustbox}
\\[2pt]
{\scriptsize Depth Pro fails the inclusion check and answers no query ray; it is reported in \cref{supp:bench}.}
\end{table}

%% file: fig/findings.tex
\begin{figure}[t]
\centering
\includegraphics[width=\columnwidth]{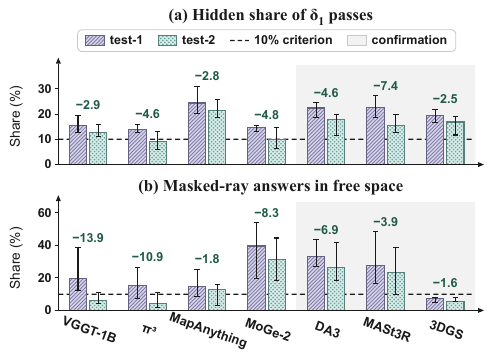}
\caption{\textbf{Registered benchmark statistics.} (a)~Share of $\delta_1$ passes on hidden-endpoint rays and (b)~share of masked-ray answers in certified free space, with 95\% window-bootstrap intervals. Above each pair: \mbox{test-2} $-$ \mbox{test-1} (percentage points). Dashed: 10\% criterion. Registered tests count the confirmation models (shaded) on \mbox{test-1} and all seven on \mbox{test-2}.}
\label{fig:findings}
\end{figure}

%% file: sec/5_method.tex
\section{Learning When to Answer}
\label{sec:method}

\input{fig/qual}

\paragraph{Problem statement.}
Selective reconstruction is useful when an application prioritizes geometry observed in the input views. Given a frozen reconstruction, a selector decides whether to return the backbone's depth or abstain on each query ray. \head{} learns from cues in the reconstruction, including coverage, point density, and predicted geometric evidence, using the training windows' endpoint labels (\cref{fig:overview}, bottom). Observed endpoints (\stV) are positive training examples, and hidden endpoints (\stH) are negative examples. At matched recall of observed endpoints, we measure how many evidence-violating answers the selector returns per 1{,}000 hidden-endpoint rays.
\vspace{-5.5pt}

\paragraph{Backbone.}
The backbone is \mbox{VGGT-1B}, frozen, which predicts cameras, point maps, and per-pixel confidence in one pass. For a query ray it supplies the readout depth of \cref{sec:benchmark}, scaled by one factor ($s = 1.0096$) from the training windows. \head either answers with this depth or abstains, so every effect it has comes from where it answers.

\subsection{\texorpdfstring{\head}{ID-Head}}
\label{sec:task}

The labels of \head are the endpoint states: \stV rays are positives and all other rays negatives, so an answer on an \stH ray stays a negative even when it is accurate. A selector trained on error would keep such answers, which the models give within $\tauv$ on 9--13\% of the \stH rays (\cref{supp:bench}).

Each ray carries 18 features computed from the backbone's predictions and cameras alone (\cref{supp:head}): readout range and coverage, confidence, point count and second z-buffer gap, two geometric terms, and the \emph{predicted evidence}, the construction of \cref{sec:evidence} run on the backbone's own input-view depths (e.g., $\hat b$, the evidence gap $\hat u/\hat L$, and whether the answer lies in $\hat\free$). A U-Net~\cite{ronneberger2015unet} with three downsamplings (0.35M parameters, receptive field about $47{\times}47$ rays) maps the $18{\times}90{\times}160$ feature map of a query view to one logit per ray. It is trained with a weighted binary cross-entropy on the full maps of 1{,}562 training queries, with stratum weights and weight 3 on \stH rays whose answer violates the evidence, using Adam~\cite{kingma2015adam} for 10 epochs and five seeds.

\subsection{Protocol}
\label{sec:protocol}

The comparators are training-free signals of the same features, \emph{chosen on each test set}, which favors them: for ranking, the best of 13 single features; for selective answering, the best of six gates (confidence, point count, evidence gap, readout range, the evidence wrapper that answers when $\hat b$ is finite, and random abstention). Ranking uses the window-median AUROC of \stV versus \stH on answered rays. Selective answering counts answered \stH rays that violate the evidence, per 1{,}000 \stH rays, averaged over \stV recalls 0.30--0.90 so that no single operating point decides the result. The ranking hypothesis requires the lower 95\% bound of the AUROC difference to exceed 0, the violation hypothesis the upper bound of the ratio to the best gate to stay below 1 (paired window bootstrap), each in at least three of five seeds. Protocol, checkpoints, and comparators were registered before any method result on either test set was computed.

\input{generated/tab4_method}
\input{fig/curves}

\subsection{Results}
\label{sec:methodresults}

\paragraph{\head answers fewer violating rays than every training-free gate.}
Both registered hypotheses hold in all five seeds on both test sets (\cref{tab:method}). On \mbox{test-1}, \head reaches AUROC 0.841 against 0.786 for the best training-free signal, the predicted evidence gap, and answers 11.67 violating rays per 1{,}000 \stH rays against 15.27 for the same gap used as a gate (ratio 0.75--0.78, upper bounds 0.88--0.91). On \mbox{test-2} the reduction is larger: 4.21 against 6.39 (ratio 0.63--0.71), with AUROC 0.887 against 0.854. The averages do not rest on a few operating points: at every recall on both test sets, every seed answers fewer violating rays than every gate (\cref{fig:curves}). Abstaining costs no accuracy: at \stV recall 0.807, $\delta_1$ on answered \stV rays is 0.979--0.980 on \mbox{test-1} and 0.985--0.986 on \mbox{test-2}, against 0.980 and 0.978 for the backbone on all \stV rays it answers.
\vspace{-5.5pt}

\paragraph{The strongest gate is an identified-set quantity.}
The backbone's confidence is a weak gate (22.6 and 21.2 violating answers against 30.2 and 23.0 for random abstention), whereas the best gate on both sets is the evidence gap, an identified-set quantity on predicted geometry. Learning from identified-set labels removes a further 22--25\% and 29--37\% of the violations. \Cref{fig:qual} shows where: the gate answers a cluster of hidden-endpoint rays that no input view saw, and \head abstains on most of it while answering as many observed-endpoint rays. Over \mbox{test-1}, the advantage lies on rays without a predicted upper bound (11.8--12.2 against 25.0 thousand violating answers at recall 0.807).

\subsection{What the head relies on}
\label{sec:analysis}

\noindent\textbf{Labels and evidence channels.}
Every learned selector of the registration is trained on the same labels, and on \mbox{test-1} each answers fewer violating rays than the best gate, with or without the ten evidence channels (\cref{supp:head}). Without them the head matches the registered one on validation and does at least as well on \mbox{test-1} (ratios 0.67--0.72), but on \mbox{test-2} it meets neither criterion (0.81--0.86, upper bounds above 1) while the registered head meets both. The predicted evidence gap is also a stronger signal there. Shifting the evidence channels along the rays of a query lowers validation AUROC from about 0.90 to 0.65--0.67, and more labels help (ratios 0.82--0.84, 0.74--0.75, 0.68--0.72 to the evidence wrapper for 25, 50, 100 training windows).
\vspace{2.5pt}

\noindent\textbf{Spatial context.}
A per-ray gradient-boosted classifier~\cite{friedman2001gbdt} on the same features reaches ratios of 0.87 [0.75, 1.004] on \mbox{test-1} and 0.99 [0.82, 1.13] on \mbox{test-2}, a gap to \head consistent with the value of spatial context.
\vspace{2.5pt}

\noindent\textbf{Transfer to another backbone.}
\head is trained for one backbone. Applied without retraining to \pithree, whose own point count is a stronger comparator (AUROC 0.807), it meets the ranking criterion in two of five seeds and the violation criterion in none (0.90--1.07), while the variant without evidence channels (0.72--0.79) and the per-ray classifier (0.86) meet the latter. Since the endpoint labels depend only on the scene and the inputs, a head for a new backbone needs only its predictions on the training windows.

%% file: fig/qual.tex
\begin{figure*}[t]
\centering
\includegraphics[width=\textwidth]{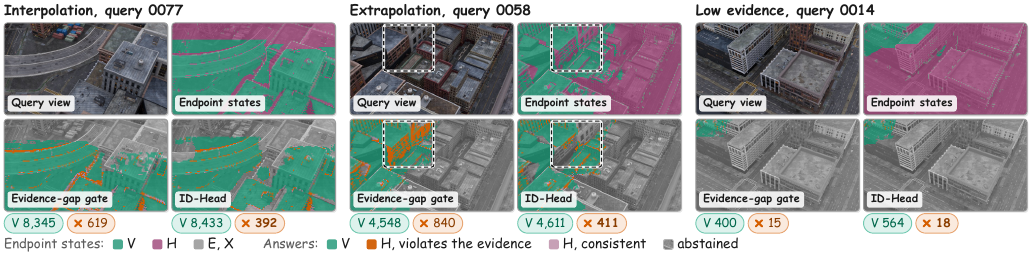}
\caption{\textbf{Where \head abstains.} Three \mbox{test-1} queries at \stV recall 0.807, one per stratum: query 0058 of \cref{fig:overview} and, in the other two strata, the query with the median change in violating answers. Untinted rays are abstentions (ties at the gate's threshold abstain); badges count answered \stV rays and violating answers. Dashed box: 573 of the gate's 840 violating answers lie inside it, against 175 of 411 for \head.}
\label{fig:qual}
\end{figure*}

%% file: generated/tab4_method.tex
\begin{table}[t]
\centering
\caption{\textbf{Registered tests of \head on VGGT-1B.} Test-1 has 10 windows, test-2 has 8. Five seeds; mean $\pm$ s.d.\ or range. Comparators: the best training-free signal and gate \emph{on each test set} (all six gates in \cref{fig:curves}). Difference: median over windows of per-window AUROC differences. Violations: violating \stH answers per 1{,}000 \stH rays, averaged over \stV recall 0.30--0.90. Green: the paired window-bootstrap bound meets the criterion in every seed. Best in red, second underlined.}
\label{tab:method}
\tabsetup
\begin{adjustbox}{max width=\columnwidth}
\begin{tabular}{?l|c|c?}
\tabtop
Ranking: AUROC of \stV vs.\ \stH $\uparrow$ & Test-1 & Test-2 \\
\tabhead
Best training-free signal ($\hat u/\hat L$) & \secondval{0.786} & \secondval{0.854} \\
\rowcolor{tabours} \head & \bestval{0.841 $\pm$ 0.005} & \bestval{0.887 $\pm$ 0.003} \\
\rowcolor{tabours} \quad difference & 0.050--0.055 & 0.025--0.033 \\
\rowcolor{tabours} \quad 95\% CI lower bound & \cellmet 0.0092--0.0247 & \cellmet 0.0035--0.0068 \\
\tabbottom
\tabgap
\tabtop
Selective answering: violations $\downarrow$ & Test-1 & Test-2 \\
\tabhead
No gate (random abstention) & 30.22 & 22.96 \\
Model confidence & 22.62 & 21.17 \\
Evidence gap $\hat u/\hat L$ & \secondval{15.27} & \secondval{6.39} \\
\rowcolor{tabours} \head & \bestval{11.67 $\pm$ 0.15} & \bestval{4.21 $\pm$ 0.20} \\
\rowcolor{tabours} \quad ratio to the best gate & 0.75--0.78 & 0.63--0.71 \\
\rowcolor{tabours} \quad 95\% CI upper bound & \cellmet 0.88--0.91 & \cellmet 0.73--0.82 \\
\tabbottom
\end{tabular}
\end{adjustbox}
\end{table}

%% file: fig/curves.tex
\begin{figure}[t]
\centering
\includegraphics[width=\columnwidth]{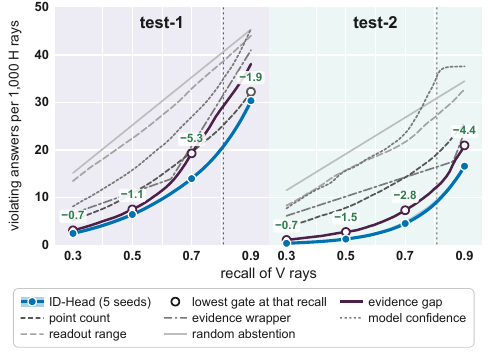}
\caption{\textbf{Violation--recall curves.} Violating answers on hidden-endpoint rays against the recall of observed-endpoint rays (lower is better). The band spans the five seeds of \head; at every recall, every seed lies below every gate. Green: \head minus the lowest gate. Dotted: recall 0.807 (\cref{fig:qual}).}
\label{fig:curves}
\end{figure}

%% file: sec/6_conclusion.tex
\section{Conclusion}
\label{sec:conclusion}

An unobserved endpoint does not make every prediction defensible. \bench{} evaluates novel-view depth against input-view geometry, exposing contradictions that depth thresholds accept and visibility masks omit. Counterfactual twins preserve per-ray evidence verdicts despite changing reference depths. Endpoint labels support selective reconstruction on a frozen VGGT-1B, reducing evidence-violating hidden-endpoint answers by 22--37\% relative to the strongest training-free gate, averaged over matched observed-endpoint recalls. These results support evaluating scene accuracy and evidence consistency as complementary properties of reconstruction.
\vspace{2.5pt}

\noindent\textbf{Limitations.}
Evaluation covers two aerial test blocks of one synthetic city with exact reference depth and poses. Real UAV evaluation requires calibrated cameras and independently measured reference geometry to assess sensitivity to depth noise, missing measurements, and pose errors.

%% file: supp/A_registration.tex
\section{Registration history and outcomes}
\label{supp:prereg}

The study was registered in three stages. Each registration is a tag of the development repository and was created before the data it governs were scored: \texttt{prereg-v1} (commit \texttt{634e7ac}) before any model was run on test-1, \texttt{prereg-v1.1} (\texttt{f8b147f}) before the confirmation models were run, and \texttt{prereg-v1.2} (\texttt{2fbd216}) before any method result on either test set and before any model was run on test-2. The benchmark results are tagged \texttt{results-v1} (\texttt{405dfbc}) and \texttt{results-v1.1} (\texttt{7653085}), and the registered method outputs are tagged \texttt{results-v2} (\texttt{df43923}, which adds the output inventory to the run commit \texttt{9b84aed}). The archived \texttt{IDSet3D} evaluator and protocol files correspond to a snapshot of the \texttt{results-v2} state without this history: \texttt{prereg/} holds the registration files, \texttt{release/} the SHA256 of every frozen result, and \texttt{release/RELEASE\_EDITS.md} the edits made for the snapshot, none of which changes a hypothesis, threshold, numerical function, or result value. \Cref{tab:supp-hypotheses} lists every registered hypothesis and its outcome.

\begin{table*}[tp]
\centering
\caption{\textbf{Registered hypotheses and outcomes.} Window medians on the stated test set. Exploration models: VGGT-1B, \pithree, MapAnything, MoGe-2. Confirmation models: DA3, MASt3R, 3DGS. Composite ranking: mean rank of $\delta_1$ on point rays (R1; evidence categories in \cref{supp:eval}) or on \stV{} rays (R$'$1) and of FV$_{\mathrm{partial}}$ (violation rates as defined in \cref{tab:supp-base}). $v_{\mathrm{rep}}$: an added copy of an input view. $v_{\mathrm{non}}$: an added view that changes the evidence of at most 1\% of the query rays. $\sigma_{\mathrm{rep}}$: spread of the predicted depth under $v_{\mathrm{rep}}$ and input-order permutations. $\Delta U$: mean change of the predicted depth after adding $v_{\mathrm{non}}$, on uncertain rays (evidence category partial or unbounded) whose evidence is unchanged. $\Delta C(v)$: fraction of such rays whose answer state (answered or not) changes after adding view $v$. Outcome: green, met; pink, not met; yellow, not computable or met only through a tie.}
\label{tab:supp-hypotheses}
\tabsetup
\setlength{\tabcolsep}{2.3pt}
\begin{adjustbox}{max width=\textwidth}
\begin{tabular}{?l|l|l|l?}
\tabtop
\multicolumn{1}{?c|}{Hypothesis} & \multicolumn{1}{c|}{Criterion} & \multicolumn{1}{c|}{Models, data} & \multicolumn{1}{c?}{Outcome} \\
\tabhead
\multicolumn{4}{?l?}{\textit{Registration v1}} \\
R1 ranking change & Kendall $\tau_b<0.6$ or top-two swap, standard vs.\ composite ranking & exploration, test-1 & \cellopen not computable: no point rays \\
R2 confidence failure & confidence AUROC (\stV{} vs.\ \stH) $<0.7$ for $\ge$50\% of models & VGGT-1B, \pithree, MapAnything, test-1 & \cellnotmet not met: 1 of 3 \ci{(0.692, 0.784, 0.723)} \\
R3 intervention failure & $\Delta U/\sigma_{\mathrm{rep}}>3$ for $\ge$2 models & VGGT-1B, MapAnything, test-1 & \cellnotmet not met: 1.05, 0.85 \\
R4 evidence violation & FV$_{\mathrm{masked}}>10\%$ for $\ge$ half of the models & exploration, test-1 & \cellnotmet not met: 0 of 4 \ci{(2.9--6.0\%)} \\
\hline
\multicolumn{4}{?l?}{\textit{Registration v1.1}} \\
R$'$1 ranking change & as R1, with \stV-ray $\delta_1$ & confirmation, test-1 & \cellopen met only through a tie, not counted \\
R$'$3 answer-state change & $\Delta C(v_{\mathrm{non}})/\Delta C(v_{\mathrm{rep}})>3$ for $\ge$2 models & DA3, MASt3R & \cellnotmet not met: 2.51, 1.50 \\
R$'$4 free-space answers & $>10\%$ for $\ge$ half of the models & confirmation, test-1 & \cellmet met: 2 of 3 \\
R$'$5 hidden share & $\ge10\%$ for $\ge$ half of the models & confirmation, test-1 & \cellmet met: 3 of 3 \\
\hline
\multicolumn{4}{?l?}{\textit{Registration v1.2}} \\
R$'$4 replication & as R$'$4 & all seven, test-2 & \cellmet met: 4 of 7 (threshold 4) \\
R$'$5 replication & as R$'$5 & all seven, test-2 & \cellmet met: 6 of 7 \\
\hline
\multicolumn{4}{?l?}{\textit{Registration v1.2, \head}} \\
Ranking & AUROC difference, CI lower bound $>0$, $\ge$3 of 5 seeds & VGGT-1B, test-1 / test-2 & \cellmet met: 5/5, 5/5 \\
Selective answering & violation ratio, CI upper bound $<1$, $\ge$3 of 5 seeds & VGGT-1B, test-1 / test-2 & \cellmet met: 5/5, 5/5 \\
\tabbottom
\end{tabular}
\end{adjustbox}
\end{table*}

\paragraph{From v1 to v1.1.}
Three of the four original v1 outcomes trace to definitions that do not fit aerial novel views. R1 compares rankings on point-identified rays, and aerial query cameras outside the input frusta leave none (\cref{sec:states}). R4 divides the number of free-space answers by all masked rays, including the rays a model leaves unanswered, so it mixes the answer rate with the violation rate: the exploration models answer 17--38\% of the masked rays and 15--40\% of these answers lie in certified free space, while R4 is 2.9--6.0\% (each range reports per-model window medians independently, so their endpoints do not multiply). R3 measures depth changes only on rays answered both before and after a non-informative view is added, so it cannot register rays that start or stop being answered. Registration v1.1 replaced these definitions with R$'$1 (\stV-ray $\delta_1$ in the composite ranking), R$'$4 (front violations as a fraction of the \emph{answered} masked rays), and R$'$3 (changes of the answer state against a control that repeats an input view), added R$'$5, and kept R2 as registered. Because v1.1 was designed after the exploration models had been scored, it was tested only on the three confirmation models, which had not yet been run, and the main text claims only the hypotheses confirmed there.

\paragraph{R2.}
R2 misses its criterion by one model among the three exploration models with usable confidence (MoGe-2 is excluded per D7). The confidence AUROC of VGGT-1B (0.692) is below 0.7 in 6 of 10 windows, \pithree is above it (0.784), and MapAnything (0.723) falls on either side depending on the aggregation. Pooling the nine windows other than the pilot window (\texttt{mc\_yaw90\_ix11\_iy7}, which registration v1 designated to be scored first after its tag, while the metrics of the other test-1 windows stayed locked) gives 2 of 3 models, and the criterion holds in 39\% of window-bootstrap resamples. We report R2 as not met, as registered.

\paragraph{R$'$1 and R$'$3.}
Registration v1.1 records the exploration results that led to these two hypotheses: R$'$1 was met on the four exploration models (Kendall $\tau_b = -0.18$, with the two top-ranked models swapped), and R$'$3 on VGGT-1B, \pithree, and MapAnything (\cref{tab:supp-intervene}). On the confirmation models, R$'$1 is met only because the composite ranks of DA3, MASt3R, and 3DGS tie, and a tie is not evidence of a ranking change. Over all seven counted models the ranking analysis is descriptive: Kendall $\tau_b$ between the standard and the composite ranking has a window median of 0.0 (95\% CI $[-0.27, 0.20]$), and the two top-ranked models differ between the two rankings in all ten test-1 windows. R$'$3 is not met: the answer-state ratio is 2.51 for DA3 and 1.50 for MASt3R, below the threshold of 3.

\paragraph{Answer-state changes for all models.}
\Cref{tab:supp-intervene} gives the quantities behind R$'$3 for all seven counted models. A non-informative view changes the answer state of 4.3--5.1\% of the uncertain rays for VGGT-1B, MapAnything, DA3, and MASt3R, 2.4\% for \pithree, and 1.5\% for 3DGS. MoGe-2 predicts each view separately, and its answer states change by less than 0.05\% in every window. The ratio to a repeated view is 4.5--5.7 for VGGT-1B, \pithree, and MapAnything. These three models had been scored before R$'$3 was defined, so their values are descriptive. Permuting the eight inputs alone, a control that v1.1 lists as reported only, changes 2.1\% of the answer states of VGGT-1B and 4.0\% of those of MapAnything, and less than 0.01\% of those of \pithree, so the answers of the first two already depend on the input order. Measured against this permutation control instead of the repeated view, the ratio of the window medians is 2.4 for VGGT-1B and 1.2 for MapAnything: merely reordering the inputs accounts for most of the change under $v_{\mathrm{non}}$ for MapAnything and roughly 40\% of it for VGGT-1B.

\input{generated/tabS_intervene}

\paragraph{From v1.1 to v1.2.}
Registration v1.2 was written after \head had been developed on training, validation, and development windows (\cref{supp:head}). It added test-2 (block~2), fixed the head architecture, its five checkpoints, the answer scale $s$, the comparators and both method hypotheses, and listed the secondary results: operating points, the mean-logit ensemble, a per-ray classifier of gradient-boosted decision trees (GBDT), the variant without evidence channels (on test-1), zero-shot transfer to \pithree, and the mechanism split. The development windows lie in block~1 and share no frame with test-1. Their results shaped two choices of v1.2. Violations are averaged over a recall range, because on these windows the confidence gate answered fewer violating rays than the head below \stV recall 0.6 and more above 0.7, so the outcome would have depended on which single operating point was chosen. Each test set also uses its own strongest training-free comparator, because the strongest gate averaged over recall differed between the validation windows (the evidence gap) and the development windows (the point count).

\paragraph{The second data source.}
The original plan also listed an asset-inspection crane sequence as a second, mesh-based data source. A rule fixed before its test results were read excluded it from all hypotheses, because the scale implied by the environment textures of that scene disagrees with its geometry by a factor of about 8.5. It is not analyzed in this paper.

\paragraph{Deviations.}
The repository records nine deviations from the registered protocol, each with a timestamp.
(D1)~The input-order permutations required alongside the repeated-view control to compute $\sigma_{\mathrm{rep}}$ in R3 were missing from the frozen run. They were run after the \texttt{prereg-v1} tag and after R3 had been seen with the repeated view alone, and they enter only R3 and the permutation control of \cref{tab:supp-intervene}.
(D2)~An early scorer called the judge without evidence and returned no front violations. The judge now refuses to run without evidence.
(D3)~R1 is recorded as not computable, because the pilot window already had no point rays; no substitute ranking was introduced.
(D4)~The evaluator was re-tagged as \texttt{eval-v1.0.3} after D2, once a regression test had reproduced every output bit for bit; v1.1 names this version.
(D5)~Under intervention conditions, feed-forward adapters align predicted cameras to the eight original input cameras only; extra or duplicated views ($v_{\mathrm{non}}$, $v_{\mathrm{rep}}$) are not used in the Umeyama alignment.
(D6)~An interim report counted \stE and \stX rays in the R4 denominator. The registered definition excludes them, and all reported values follow it.
(D7)~The confidence output of MoGe-2 is a constant validity mask, and Depth Pro and 3DGS have no usable confidence, so their confidence AUROC and R2 entries are N/A.
(D8)~The first 3DGS test run used random initialization and failed the inclusion check on the development windows (input-view $\delta_1 = 0.14$). It was voided before any query metric was read and rerun once with COLMAP initialization from the input images and poses, keeping the number of steps, the seed, and the readout.
(D9)~The frozen evaluator computes confidence AUROC with point, supported, and free rays as the positive class, and aerial views have none of these. The confidence AUROC of R2 and of \cref{tab:supp-base} uses the \stV-versus-\stH definition of the metric specification instead; the frozen evaluator is unchanged.

\paragraph{Run repairs.}
The registered method run stopped twice before computing any metric: a test-1 window loader lacked a field that validation windows carry, and two test-1 windows staged the same depth file from different paths. Both fixes change data loading and run orchestration only (tags \texttt{prereg-v1.2-repair1}, \texttt{4934e10}, and \texttt{prereg-v1.2-repair2}, \texttt{8872a83}). No feature, label, judge, scoring function, or checkpoint changed, and the outputs of the earlier stages were reused after verification against the first run's manifest.

\paragraph{Recomputation.}
A second implementation, written separately from the formal code, recomputed the method results on test-1, test-2, and \pithree from the saved model predictions. It recomputed the readout-range, confidence, and wrapper features from the raw predictions (no difference to the stored ones), rebuilt the labels with the frozen judge, and reimplemented the AUROCs, thresholds and ties, bootstrap, and aggregates; it shares the evidence construction, the judge, the feature definitions, the checkpoints, and the model predictions with the formal run. In double precision, AUROCs, ranking differences, violation ratios, and their intervals agree with the formal outputs to $4.4\times10^{-16}$, mean violation curves to $1.5\times10^{-14}$, violation counts at P1 and P2 to $2\times10^{-11}$ (exactly for the heads), and recalls at P3 and $\delta_1$ at P2 to $10^{-6}$; the test-2 replication statistics match exactly. The P3 counts of the gates were not compared; \cref{tab:supp-operating} takes them from the formal outputs.

\paragraph{Bootstrap stream.}
The registration fixed the bootstrap seed and the number of resamples but not how random draws map to windows. The frozen code draws one $2{,}000 \times n$ index matrix over windows in sorted order. An equally valid stream that draws each resample separately moves the interval bounds by up to 0.018 for the ranking difference and 0.013 for the violation ratio. It changes three secondary verdicts (\cref{supp:head}, \cref{tab:variants,tab:supp-pi3}) and no registered hypothesis: under the alternative stream, the ranking criterion changes from pass to fail for \pithree seed 20261027 (CI lower bound $+0.0002$ with the frozen stream vs.\ $-0.0052$ with the alternative) and for the \pithree mean-logit ensemble ($+0.0012$ vs.\ $-0.0083$), whereas the violation criterion for the test-1 per-ray GBDT flips from fail to pass (CI upper bound $1.0039$ vs.\ $0.9995$). Under the alternative stream, the zero-shot \pithree head meets the ranking criterion in one of five seeds instead of two. All reported values use the frozen code.

%% file: generated/tabS_intervene.tex
\begin{table}[t]
\centering
\caption{\textbf{Answer-state changes under interventions.} Fraction of uncertain query rays with unchanged evidence whose answer state (answered or not) changes, in \%, on test-1 (window medians): after adding a non-informative view $v_{\mathrm{non}}$, after repeating an input view $v_{\mathrm{rep}}$, and after permuting the input order. \emph{Ratio}: window median of $\Delta C(v_{\mathrm{non}})/\Delta C(v_{\mathrm{rep}})$, the statistic of R$'$3, which was registered for DA3 and MASt3R. For MoGe-2 the repeated view changes no answer state in any window, so its ratio is undefined, and for 3DGS it is defined in nine of the ten windows; 3DGS was not run with permuted inputs. Gray: confirmation models.}
\label{tab:supp-intervene}
\tabsetup
\begin{adjustbox}{max width=\columnwidth, gstore width=\tabwidth}
\begin{tabular}{?r|C{24pt}C{24pt}C{24pt}|C{70pt}?}
\tabtop
\tabfirst{2}{Model} & \multicolumn{3}{c|}{Changed answer states (\%)} & R$'$3 statistic \\
\cline{2-5}
 & $v_{\mathrm{non}}$ & $v_{\mathrm{rep}}$ & perm. & Ratio \ci{[95\% CI]} \\
\tabhead
VGGT-1B~\cite{wang2025vggt} & 4.91 & 0.87 & 2.05 & 5.57 \ci{[4.32, 6.61]} \\
\pithree~\cite{wang2025pi3} & 2.36 & 0.55 & $<$0.01 & 4.54 \ci{[2.08, 8.82]} \\
MapAnything~\cite{keetha2025mapanything} & 4.74 & 0.81 & 4.02 & 5.65 \ci{[2.63, 8.79]} \\
MoGe-2~\cite{wang2025moge2} & 0.00 & 0.00 & 0.00 & \emptycell \\
\rowcolor{tabgroup} DA3~\cite{lin2025da3} & 4.33 & 1.74 & 0.40 & \cellnotmet 2.51 \ci{[1.71, 4.66]} \\
\rowcolor{tabgroup} MASt3R~\cite{leroy2024mast3r} & 5.14 & 3.93 & 1.79 & \cellnotmet 1.50 \ci{[0.73, 2.59]} \\
\rowcolor{tabgroup} 3DGS~\cite{kerbl20233dgs} & 1.45 & 1.16 & N/A & 0.56 \ci{[0.02, 1.64]} \\
\tabbottom
\end{tabular}
\end{adjustbox}
\tabnotes{Pink: below the threshold of 3 that R$'$3 registered for DA3 and MASt3R (\cref{tab:supp-hypotheses}); the other ratios are descriptive.}
\end{table}

%% file: supp/B_evaluation.tex
\section{Evaluation details}
\label{supp:eval}

\paragraph{Evidence.}
Input depth maps are rendered at $1920{\times}1080$. The observed surface $\surf$ is represented by a triangle mesh per input view: vertices are back-projected from the depth of one pixel per $4{\times}4$ block, the pixel at offset $(2, 2)$ next to the block center (a $480{\times}270$ evidence grid), and a triangle is kept if the relative depth spread of its vertices is at most 2\%, its normal makes at most $75^\circ$ with the viewing ray, and it does not touch the outermost ring of the grid. A sampled pixel that is edge-ambiguous in its own depth map, by the criterion given below for query pixels, does not become a vertex. The free-space test uses the minimum native depth within each block, followed by a $3{\times}3$ minimum filter on the block grid, and pixels without a hit take the range $T$. Ray sampling uses the center of the sampled native pixel, never the center of an evidence block, which avoids a half-pixel bias for even block sizes.

\begin{figure*}[tp]
\centering
\includegraphics[width=\textwidth]{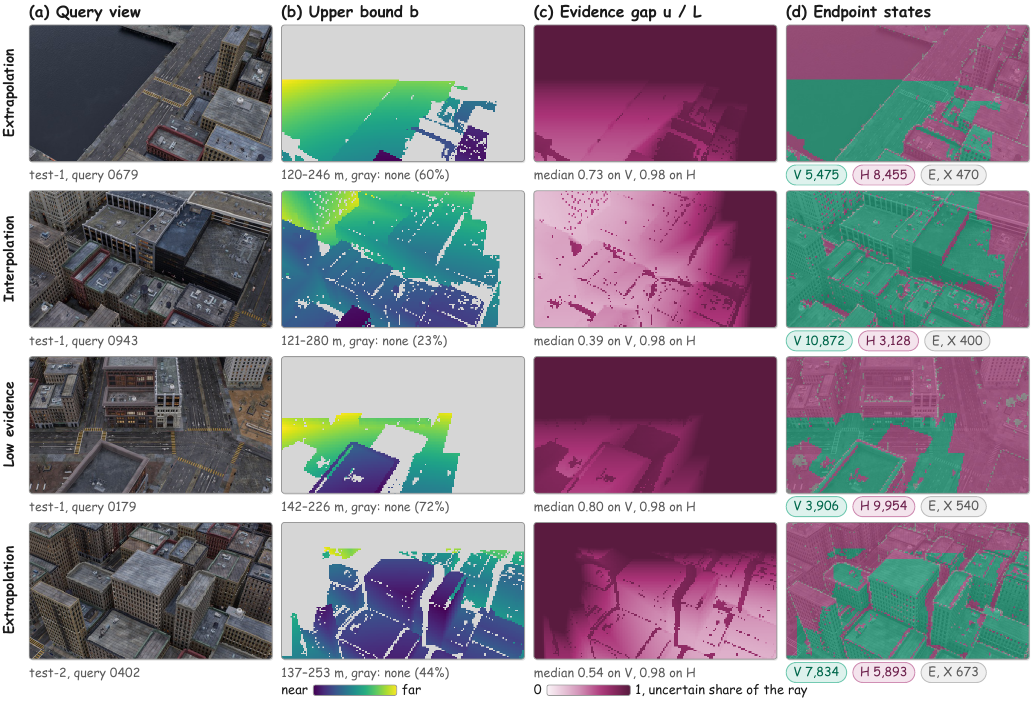}
\caption{\textbf{From evidence to endpoint states} on four queries, one per row: test-1 queries 0679, 0943, and 0179 and test-2 query 0402, labeled by stratum. (a)~Query view. (b)~Upper bound $b$, the first observed surface on the ray; gray rays meet none. (c)~Evidence gap $u/L$. (d)~Endpoint states: a scored ray is \stV if its true first hit lies on the observed surface at $b$, and \stH otherwise; \stE and \stX rays are excluded. Below each row: the range of $b$ with the share of rays without a bound, the median gap on \stV and \stH rays, and the number of rays in each state.}
\label{fig:supp-evidence}
\end{figure*}

\paragraph{Query rays and tolerances.}
Each query view is evaluated on a $90{\times}160$ grid of native pixels with a stride of 12. Along each ray, $N = 256$ midpoint samples on $[0, L]$ determine the uncertain length $u$, and the tolerance is $\tauv(d) = 0.05\,\mathrm{m} + 0.01\,d + 2L/N$. Front violations test membership in $\free$ using the per-input-view margin $\tau_s(z)$ from \cref{eq:surface}. Back violations use the ray-wise threshold $b + \tauv(b)$, and the endpoint labels use $\tauv(b)$ and the free-space test. The sensitivity analysis in \cref{supp:bench} scales both tolerances by factors of 0.5 and 2.

\paragraph{Evidence categories.}
The evaluator also assigns every ray one of five evidence categories. A ray is \emph{point} if $b$ is finite and $u \le \tauv(L)$, so that the evidence fixes its depth; \emph{free} if $b = +\infty$ and $u \le \tauv(L)$; \emph{supported} if $u > \tauv(L)$ but the evidence gap $u/L$ is at most 0.1; and \emph{partial} or \emph{unbounded} if the gap exceeds 0.1, depending on whether $b$ is finite. Hypothesis R1, deviation D9, and the features of \head refer to these categories.

\paragraph{Edge-ambiguous and inconsistent rays.}
Rendered depth mixes foreground and background at silhouettes. A query pixel is edge-ambiguous (\stE) if a pixel within Chebyshev radius 2 differs in depth by at least 2\%, where invalid depths count as the far plane. \stE is defined from the query depth map alone and reads neither the evidence nor any model output. A finite true hit that lies more than $\tauv(b)$ behind a finite bound, or in the interior of $\free$, is flagged as evidence-inconsistent (\stX; Sec.~3.3 of the main paper). Together, \stE and \stX rays make up 3.5\% of the test-1 rays and 2.7\% of the test-2 rays, and both are excluded from every metric except the twin scores, which use every ray of the divergence region $D$.

\paragraph{Examples.}
\Cref{fig:supp-evidence} shows these quantities on four queries. On the extrapolation query in the first row, 5{,}693 of the 14{,}400 rays intersect an observed surface and have a finite upper bound $b$ (120--246\,m). Of the 14{,}400 rays, the true first hit lies on that surface for 5{,}475 (\stV) and is hidden for 8{,}455 (\stH), 117 of which hit a surface in front of $b$; the remaining 470 are excluded as \stE or \stX, including the other 101 rays with a finite $b$. Over the four queries, the share of rays without a bound ranges from 23\% on the interpolation query to 72\% on the low-evidence query, and the median evidence gap on \stV rays grows with it, from 0.39 to 0.80, while it stays at 0.98 on \stH rays.

\paragraph{Readout.}
Feed-forward models predict a point map for each input view. We align the predicted input cameras to the true input cameras with a similarity transform estimated by Umeyama's method~\cite{umeyama1991} on the camera centers and the endpoints of the camera axes, using the eight input cameras only. The registered fallback for a feed-forward model without predicted cameras scales its depth by the median ratio of true to predicted depth on the input views. Monocular models (MoGe-2, Depth Pro) instead keep their metric depth and use the true input cameras, as registered. The aligned points of all input views are splatted into the query camera with a z-buffer at $4{\times}$ supersampling, and a pixel that receives no point answers $+\infty$. 3DGS is optimized for 3{,}000 steps per window on the input images with true poses and COLMAP initialization, and its rendered depth is read at the query camera, with accumulated opacity below 0.5 read as $+\infty$.

\paragraph{Inclusion check.}
A model is counted in the registered tests if, on the two development windows, its input-view $\delta_1$ is at least 0.80 and its input-view coverage at least 80\%. The thresholds are loose by design: they detect a broken pipeline or data error, not a domain gap. Depth Pro fails the check: on a development image its median metric depth is 3.2\,m with the true focal length and 2.8\,m with its own estimate, against a true median of 165\,m.

\paragraph{Twins.}
Box centers are proposed on query rays inside the uncertain part of $\idset$ in front of the true hit. A proposal is accepted only if it passes the five checks of \cref{sec:twins}. Condition~(i) requires every input pixel ray, at native resolution and on the $480{\times}270$ evidence grid, to miss the box or to enter it no earlier than $\tauv$ before the pixel's recorded depth. Condition~(iii) is checked against every fourth frame of the city block, with the same tolerance. Each test-1 window contributes four accepted twins. For every twin and model, the single prediction is scored against the query ground truth of both worlds on the divergence region $D$.

\paragraph{Statistics.}
The window is the unit of resampling. Intervals are percentile intervals from 2{,}000 window-bootstrap resamples with seed 20261017. The method comparisons resample windows jointly for both arms of a comparison and keep thresholds fixed at their full-set values.

%% file: supp/C_benchmark.tex
\section{Additional benchmark results}
\label{supp:bench}

\begin{table*}[tp]
\centering
\caption{\textbf{Metrics on test-1.} Window medians (upper box) with 95\% window-bootstrap intervals (lower box). Gray: confirmation models. Columns with an arrow: best of the seven counted models in red, second underlined.}
\label{tab:supp-base}
\tabsetup
\setlength{\tabcolsep}{4.6pt}
\begin{adjustbox}{max width=\textwidth, gstore width=\tabwidth}
\begin{tabular}{?r|cc|ccc|c|cc?}
\tabtop
\tabfirst{2}{Model} & \multicolumn{2}{c|}{Standard} & \multicolumn{3}{c|}{Identified-set violations} & Confidence & \multirow{2}{*}{UCR} & \multirow{2}{*}{\stH{} within $\tauv$} \\
\cline{2-7}
\multicolumn{1}{?c|}{} & $\delta_1$ $\uparrow$ & AbsRel $\downarrow$ & FV$_{\mathrm{masked}}$ $\downarrow$ & FV$_{\mathrm{partial}}$ $\downarrow$ & BV$_{\mathrm{partial}}$ $\downarrow$ & AUROC $\uparrow$ & & \\
\tabhead
VGGT-1B~\cite{wang2025vggt}\venue{CVPR 2025} & 0.558 & \secondval{0.039} & \secondval{0.041} & \secondval{0.244} & 0.159 & 0.692 & 0.317 & 0.126 \\
\pithree~\cite{wang2025pi3}\venue{ICLR 2026} & 0.557 & \bestval{0.025} & \bestval{0.029} & \bestval{0.036} & \bestval{0.051} & \bestval{0.784} & 0.447 & 0.134 \\
MapAnything~\cite{keetha2025mapanything}\venue{3DV 2026} & 0.566 & 0.098 & 0.049 & 0.421 & 0.400 & 0.723 & 0.092 & 0.090 \\
MoGe-2~\cite{wang2025moge2}\venue{NeurIPS 2025} & 0.532 & 0.052 & 0.060 & 0.582 & 0.153 & N/A & 0.161 & 0.092 \\
\rowcolor{tabgroup} DA3~\cite{lin2025da3}\venue{ICLR 2026} & \bestval{0.603} & 0.066 & 0.100 & 0.850 & \secondval{0.089} & \secondval{0.762} & 0.090 & 0.111 \\
\rowcolor{tabgroup} MASt3R~\cite{leroy2024mast3r}\venue{ECCV 2024} & \secondval{0.600} & 0.085 & 0.096 & 0.666 & 0.175 & 0.536 & 0.095 & 0.101 \\
\rowcolor{tabgroup} 3DGS~\cite{kerbl20233dgs}\venue{TOG 2023} & 0.542 & 0.836 & 0.072 & 0.526 & 0.154 & N/A & 0.248 & 0.101 \\
\hline
Depth Pro$^\dagger$~\cite{bochkovskii2025depthpro}\venue{ICLR 2025} & 0.000 & \emptycell & 0.000 & 0.000 & 1.000 & N/A & 0.000 & 0.000 \\
\tabbottom
\tabgap
\tabtop
\multicolumn{1}{?c|}{95\% interval} & $\delta_1$ & AbsRel & FV$_{\mathrm{masked}}$ & FV$_{\mathrm{partial}}$ & BV$_{\mathrm{partial}}$ & AUROC & \multicolumn{2}{c?}{\stH{} within $\tauv$} \\
\tabhead
VGGT-1B & \ci{[0.516, 0.599]} & \ci{[0.031, 0.045]} & \ci{[0.029, 0.085]} & \ci{[0.073, 0.683]} & \ci{[0.042, 0.319]} & \ci{[0.679, 0.743]} & \multicolumn{2}{c?}{\ci{[0.101, 0.146]}} \\
\pithree & \ci{[0.520, 0.610]} & \ci{[0.021, 0.034]} & \ci{[0.011, 0.045]} & \ci{[0.017, 0.138]} & \ci{[0.027, 0.371]} & \ci{[0.761, 0.832]} & \multicolumn{2}{c?}{\ci{[0.100, 0.154]}} \\
MapAnything & \ci{[0.521, 0.618]} & \ci{[0.086, 0.128]} & \ci{[0.034, 0.081]} & \ci{[0.284, 0.701]} & \ci{[0.186, 0.620]} & \ci{[0.683, 0.735]} & \multicolumn{2}{c?}{\ci{[0.053, 0.154]}} \\
MoGe-2 & \ci{[0.513, 0.606]} & \ci{[0.046, 0.063]} & \ci{[0.038, 0.095]} & \ci{[0.288, 0.712]} & \ci{[0.065, 0.482]} & N/A & \multicolumn{2}{c?}{\ci{[0.067, 0.115]}} \\
\rowcolor{tabgroup} DA3 & \ci{[0.565, 0.693]} & \ci{[0.057, 0.070]} & \ci{[0.078, 0.147]} & \ci{[0.768, 0.888]} & \ci{[0.068, 0.110]} & \ci{[0.728, 0.779]} & \multicolumn{2}{c?}{\ci{[0.093, 0.155]}} \\
\rowcolor{tabgroup} MASt3R & \ci{[0.576, 0.629]} & \ci{[0.078, 0.093]} & \ci{[0.052, 0.138]} & \ci{[0.329, 0.884]} & \ci{[0.104, 0.555]} & \ci{[0.497, 0.563]} & \multicolumn{2}{c?}{\ci{[0.065, 0.125]}} \\
\rowcolor{tabgroup} 3DGS & \ci{[0.514, 0.604]} & \ci{[0.675, 1.117]} & \ci{[0.050, 0.082]} & \ci{[0.457, 0.592]} & \ci{[0.124, 0.202]} & N/A & \multicolumn{2}{c?}{\ci{[0.086, 0.123]}} \\
\tabbottom
\end{tabular}
\end{adjustbox}
\tabnotes{$^\dagger$Fails the inclusion check. It answers no query ray, so every ray with a finite upper bound is a back violation.\par
The standard metrics use the scored rays of a query (\stE{} and \stX{} excluded) and take the median over the queries of a window; the registered visibility-mask AbsRel returns no value in the frozen evaluator on these views and is omitted. UCR, the registered unfounded completion reward, is the fraction of partial and unbounded rays answered within $\tauv$ of the true depth; on aerial views every scored ray is partial or unbounded, so UCR also counts accurate answers on \stV{} rays. \emph{\stH{} within $\tauv$}, added after registration, restricts the same count to \stH{} rays. FV$_{\mathrm{partial}}$ and BV$_{\mathrm{partial}}$ are the front- and back-violation rates on rays with a finite upper bound, 97\% of which are observed-endpoint rays on test-1. With a tolerance of about 2\% of depth, they also register small depth biases in front of or behind observed surfaces. FV$_{\mathrm{masked}}$ divides front violations on masked rays by all masked rays, answered or not. Confidence AUROC separates \stV{} from \stH{} rays (deviation D9), and N/A marks models without a usable confidence.}
\end{table*}

\input{generated/tabS_reference}

\input{generated/tabS_twins}

\begin{figure}[tp]
\centering
\includegraphics[width=\columnwidth]{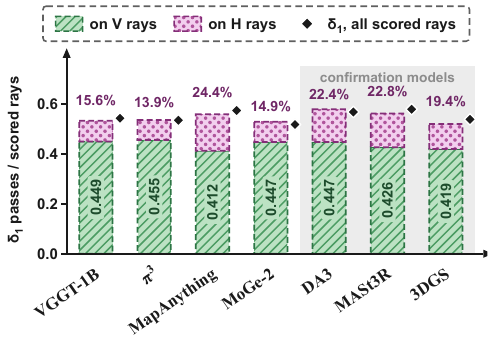}
\caption{\textbf{Where $\delta_1$ passes come from.} $\delta_1$ passes on \stV and on \stH rays as fractions of the scored rays of a window (test-1 window medians). Above each bar: the hidden share of \cref{tab:findings}. Diamond: window median of $\delta_1$ on the scored rays. The decomposition is exact within a window; the medians of the two parts need not add up to the median of their sum. Inside each bar: the \stV part. Shaded: confirmation models.}
\label{fig:supp-split}
\end{figure}

\paragraph{Two views of $\delta_1$.}
\Cref{tab:supp-base} reports $\delta_1$ as registered: per query on its scored rays, then as the median over the queries of a window. \cref{tab:findings} pools the passes of all scored rays of a window instead, because that version decomposes exactly into contributions from \stV and \stH rays. Both exclude \stE and \stX rays. The two versions order the models differently, and in both of them the spread across the counted models (0.071 and 0.061) is smaller than the credit that each model collects on hidden endpoints (0.080--0.146). \Cref{fig:supp-split} shows this decomposition. Passes on \stV rays amount to 0.41--0.46 of the scored rays for every model, and $\delta_1$ is 0.90--0.98 on \stV rays against 0.15--0.28 on \stH rays.

\paragraph{Reference answers.}
\Cref{tab:supp-reference} scores four constructed answers with the same evaluator. A single depth per query, the median true depth of its \stV rays, reaches $\delta_1 = 0.758$, above every model in \cref{tab:supp-base} (0.532--0.603), which shows how loose $\delta_1$ is at these ranges. The identified-set judgment rejects the same answer: 25.8\% of the scored rays are front violations under it. Answering the upper bound $b$ wherever it is finite yields a front-violation rate of 0.000 and leaves 51.8\% of the scored rays unanswered; because $\delta_1$ counts an unanswered ray as a failure, it scores 0.460. We also evaluate the training-free evidence wrapper in depth-substitution mode (\cref{tab:supp-reference}), which outputs the predicted bound $\hat b$ where it is finite and falls back to the model's readout elsewhere. This variant reaches $\delta_1 = 0.563$ with either backbone and leaves 42.0\% (VGGT-1B) and 42.5\% (\pithree) of the scored rays unanswered (rays where both $\hat b = +\infty$ and the backbone produces no splatted depth). In contrast, when used as a selective-answering gate (Sec.~5.2 of the main paper and \cref{tab:supp-operating,tab:supp-pi3}), the same finite-$\hat b$ condition only decides whether the backbone's depth is answered.

\paragraph{Twins.}
\Cref{tab:twins} reports the per-model scores on the divergence regions of the 40 twins summarized in \cref{sec:findings}. The example of \cref{fig:teaser} is query~0687 of the test-1 window \texttt{mc\_yaw180\_ix13\_iy3} with its twin~1, a box of $4.6\times14.6\times7.8$\,m. The box hides 1{,}743 of the 14{,}400 rays of the query, and VGGT-1B answers 1{,}730 of them; $\delta_1$ on all 1{,}743 rays, an unanswered ray counting as a failure, is 0.984 against world~A (0.991 on the answered rays) and 0.000 against world~B. On the same rays the evidence check judges 664 rays consistent with the inputs and 1{,}079 contradicted by them (1{,}049 front violations and 39 back violations, with 9 rays incurring both) in both worlds; Fig.~1 draws one dot per ray. The city of Fig.~1 is built from the ground-truth depth of the window's frames in block~1; a surface keeps its color when one of the eight input views records it within 0.05\,m plus 1\% of its depth at a pixel of the $3{\times}3$ neighborhood of its projection, a slightly looser form of the observed-surface test of Sec.~3.2 of the main paper. The 26\% in the caption is the share of hidden-endpoint rays among the scored rays of query~0687 (3{,}562 of 13{,}700).

\input{generated/tabS_strata}

\begin{figure*}[tp]
\centering
\includegraphics[width=\textwidth]{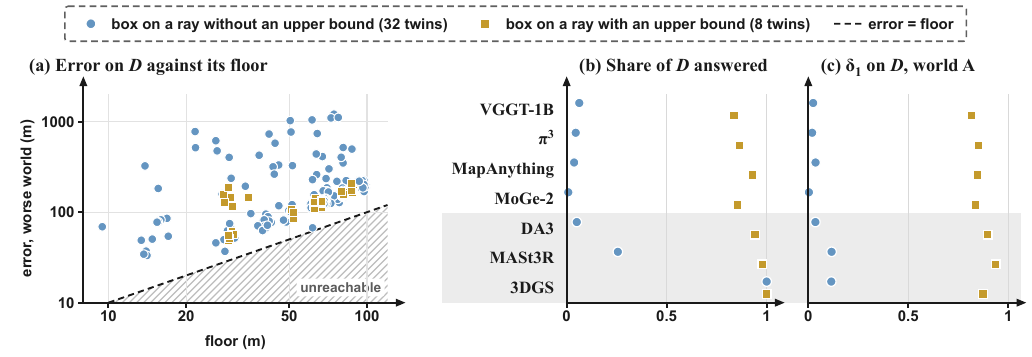}
\caption{\textbf{Twins.} (a)~Error in the worse of the two worlds, averaged over the answered rays of $D$, against its floor; one point per twin and model with answered rays (logarithmic axes). (b)~Fraction of $D$ answered and (c)~$\delta_1$ on $D$ against world~A, as medians over the twins of each box type. Shaded: confirmation models.}
\label{fig:supp-twins}
\end{figure*}

\paragraph{Twins by box type.}
The 40 boxes are of two types, by the query ray on which the box was proposed (\cref{fig:supp-twins}b, c). For 32 twins this ray has no upper bound. The six learned models (all counted models except 3DGS) answer a median of 1--26\% of the divergence region there, and their median $\delta_1$ in world~A is at most 0.12. For the other 8 twins the box was proposed in front of an observed surface. The learned models answer a median of 84--98\% of these regions and reach a median $\delta_1$ of 0.82--0.94 in world~A and 0.000 in world~B. The values of 0.22--0.33 in \cref{tab:twins} are medians over both types and are typical of neither. \Cref{fig:supp-twins}a plots the error of every model on every twin against its floor: over the 200 pairs of twin and model with answered rays, the median error in the worse world is 2.0 times its floor.

\begin{figure*}[tp]
\centering
\includegraphics[width=\textwidth]{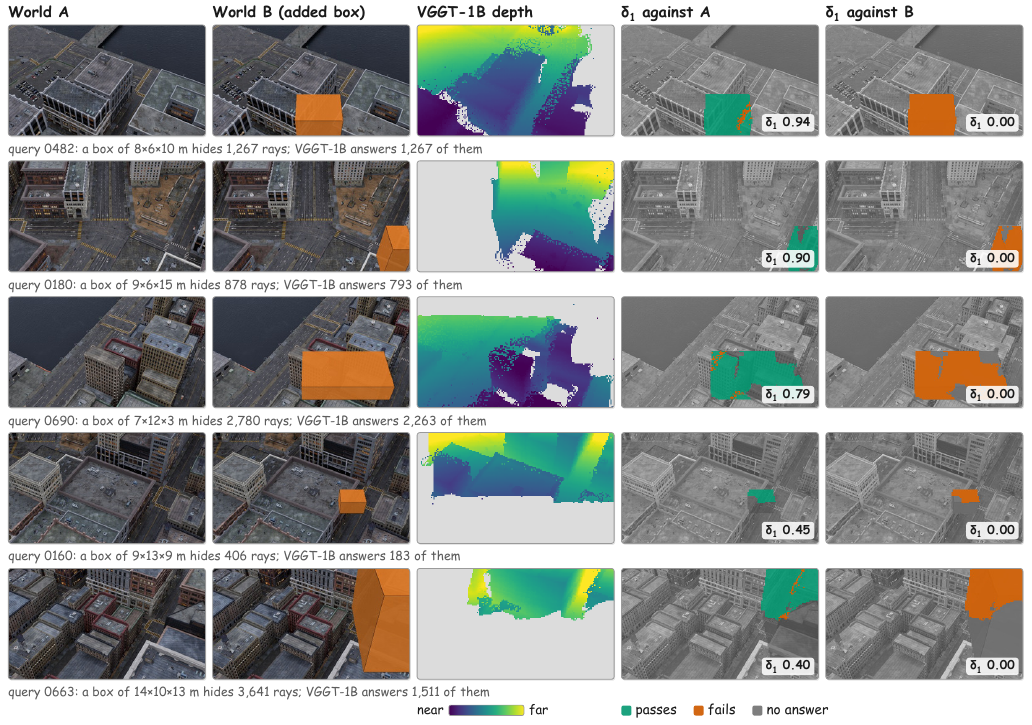}
\caption{\textbf{Five twins on test-1 queries.} From left: world~A, world~B with the added box (orange), the depth predicted by VGGT-1B, which is the same for both worlds, and the judgment of $\delta_1$ on the rays the box hides against each world. Below each row: the box size, the rays it hides, and how many of them VGGT-1B answers. Badges: $\delta_1$ over these rays, with unanswered rays counted as failures. Light gray in the depth column: no answer.}
\label{fig:supp-twins-real}
\end{figure*}

\paragraph{Twins on single queries.}
\Cref{fig:supp-twins-real} shows five twins, each on one query of its window. In the first three, the box was proposed in front of an observed surface. It hides 878--2{,}780 rays of the query, VGGT-1B answers 81--100\% of them, and $\delta_1$ is 0.79--0.94 against world~A. The last two boxes were proposed on rays without an upper bound; VGGT-1B answers 45\% and 41\% of the rays they hide, and $\delta_1$ against world~A is 0.45 and 0.40. Against world~B, no answer on these rays passes $\delta_1$ in any of the five twins.

\begin{figure*}[tp]
\centering
\includegraphics[width=\textwidth]{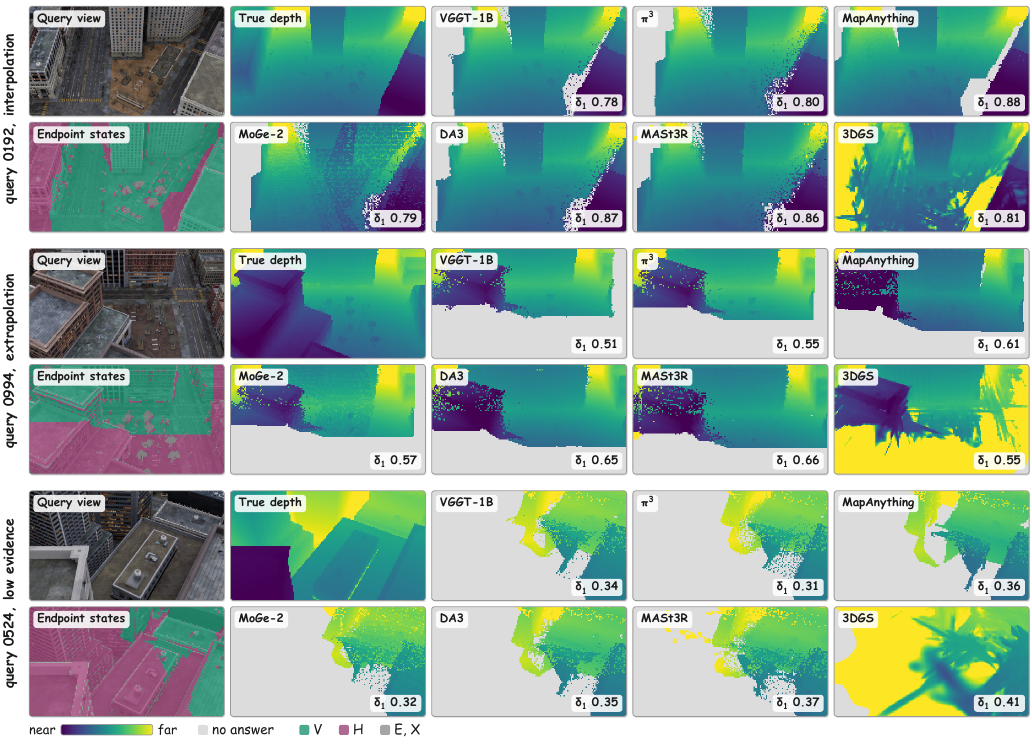}
\caption{\textbf{Depth of the seven counted models} on one test-1 query of each stratum, each from a different window, with the query view, the true depth, and the endpoint states. One color scale per query; gray: no answer. Badges: $\delta_1$ over the scored rays, with unanswered rays counted as failures.}
\label{fig:supp-depth}
\end{figure*}

\paragraph{Query strata.}
\Cref{tab:supp-strata} splits the metrics of \cref{tab:supp-base} by query stratum. For every model, $\delta_1$ falls from 0.76--0.83 on interpolation queries to 0.54--0.60 on extrapolation queries and 0.26--0.32 on low-evidence queries, and FV$_{\mathrm{masked}}$ falls from 0.05--0.20 to 0.03--0.11 and 0.01--0.03. FV$_{\mathrm{masked}}$ divides by all masked rays, answered or not, so it also falls when fewer masked rays are answered and does not measure how often an answer violates the evidence. FV$_{\mathrm{partial}}$ shows no consistent trend across strata. A window holds 5--8 interpolation, 8 extrapolation, and 2--7 low-evidence queries. \Cref{fig:supp-depth} shows the depths of the seven counted models on one query of each stratum. Their $\delta_1$ is 0.78--0.88 on the interpolation query, 0.51--0.66 on the extrapolation query, and 0.31--0.41 on the low-evidence query. On the last, the six learned models answer 36--43\% of the scored rays, and 3DGS, which answers 93\%, has the highest $\delta_1$.

\begin{figure*}[tp]
\centering
\includegraphics[width=\textwidth]{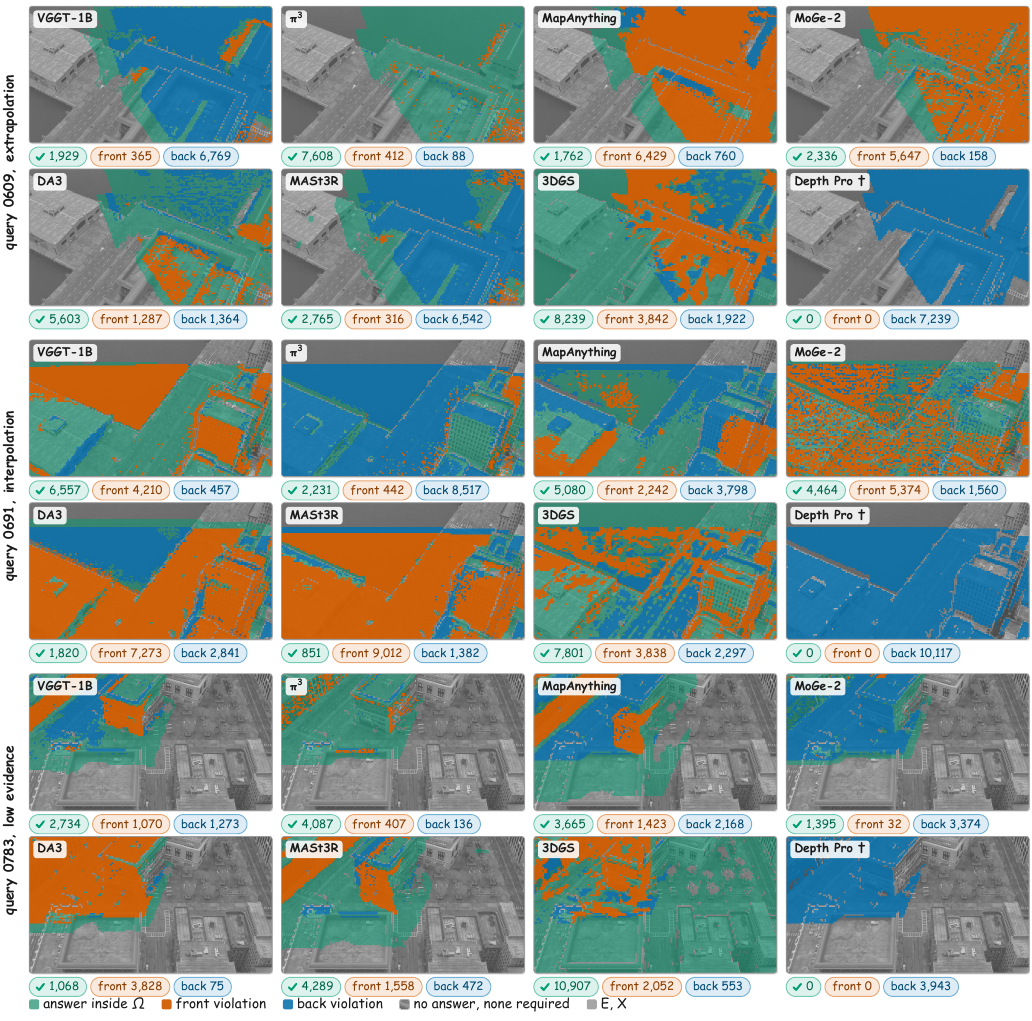}
\caption{\textbf{Identified-set judgment of eight models on three test-1 queries} (frozen evaluator): queries 0609, 0691, and 0783, one per stratum and each from a different window. Badges count answers inside $\idset$, front violations, and back violations among the scored rays. A back violation is an answer behind $b + \tauv(b)$, or no answer on a ray with a finite $b$; a ray flagged both ways is drawn and counted as a front violation. $^\dagger$Depth Pro fails the inclusion check and answers no ray.}
\label{fig:supp-judge}
\end{figure*}

\input{generated/tabS_base_t2}

\paragraph{Per-ray judgments.}
\Cref{fig:supp-judge} shows the judgment of every model's answers on one query of each stratum. On the extrapolation query (first row; 5{,}218 \stV and 8{,}785 \stH rays), MapAnything and MoGe-2 place 6{,}429 and 5{,}647 answers in certified free space, VGGT-1B and MASt3R have 6{,}769 and 6{,}542 back violations, and \pithree has 412 front and 88 back violations. Most of these violations lie on \stV rays (4{,}293 of 6{,}429 for MapAnything; 4{,}789 of 6{,}769 for VGGT-1B): they reflect depth biases exceeding the tolerance $\tauv$ in front of or behind observed surfaces (as noted in \cref{tab:supp-base}), rather than free-space answers on \stH rays. On the \stH rays, MapAnything and MoGe-2 have 2{,}136 and 1{,}806 front violations, and \pithree has 291. The other two queries show the same pattern: 8{,}437 of the 9{,}012 front violations of MASt3R on the interpolation query and 3{,}401 of the 3{,}828 of DA3 on the low-evidence query lie on \stV rays, and the \stH rays of these queries carry 12--668 front violations per counted model.

\begin{figure*}[tp]
\centering
\includegraphics[width=\textwidth]{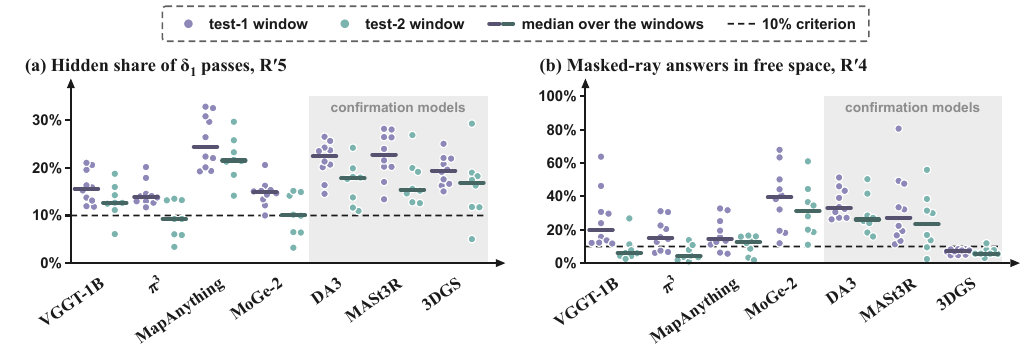}
\caption{\textbf{The registered statistics window by window.} One point per window (10 on test-1, 8 on test-2). Horizontal line: the median over the windows, the value in \cref{tab:findings}. Dashed: 10\% criterion. Shaded: confirmation models.}
\label{fig:supp-windows}
\end{figure*}

\begin{table}[tp]
\centering
\caption{\textbf{Sensitivity to the tolerances.} Multipliers apply jointly to $\tau_s$ and $\tauv$ (values in \%). Blue: the registered setting.}
\label{tab:supp-tau}
\tabsetup
\begin{adjustbox}{max width=\columnwidth}
\begin{tabular}{?r|C{21pt}C{21pt}C{21pt}|C{21pt}C{21pt}C{21pt}?}
\tabtop
\tabfirst{2}{Model} & \multicolumn{3}{c|}{Free-space answers} & \multicolumn{3}{c?}{Hidden share} \\
\cline{2-7}
 & $0.5\times$ & \cellcolor{tabours}$1\times$ & $2\times$ & $0.5\times$ & \cellcolor{tabours}$1\times$ & $2\times$ \\
\tabhead
VGGT-1B~\cite{wang2025vggt} & 20.5 & \cellcolor{tabours}19.9 & 17.3 & 15.8 & \cellcolor{tabours}15.6 & 15.5 \\
\pithree~\cite{wang2025pi3} & 16.2 & \cellcolor{tabours}15.0 & 12.8 & 14.2 & \cellcolor{tabours}13.9 & 13.8 \\
MapAnything~\cite{keetha2025mapanything} & 15.1 & \cellcolor{tabours}14.5 & 13.7 & 24.7 & \cellcolor{tabours}24.4 & 24.3 \\
MoGe-2~\cite{wang2025moge2} & 40.5 & \cellcolor{tabours}39.6 & 32.0 & 15.0 & \cellcolor{tabours}14.9 & 14.8 \\
DA3~\cite{lin2025da3} & 34.9 & \cellcolor{tabours}33.1 & 31.4 & 22.7 & \cellcolor{tabours}22.4 & 22.3 \\
MASt3R~\cite{leroy2024mast3r} & 28.1 & \cellcolor{tabours}27.3 & 26.0 & 23.0 & \cellcolor{tabours}22.8 & 22.7 \\
3DGS~\cite{kerbl20233dgs} & 7.9 & \cellcolor{tabours}7.3 & 6.0 & 19.6 & \cellcolor{tabours}19.4 & 19.3 \\
\tabbottom
\end{tabular}
\end{adjustbox}
\end{table}

\paragraph{Replication on test-2.}
On the eight test-2 windows, the free-space answer rate exceeds 10\% for MoGe-2 (31.3\%), DA3 (26.2\%), MASt3R (23.4\%), and MapAnything (12.7\%), and stays below it for VGGT-1B (6.0\%), \pithree (4.1\%), and 3DGS (5.7\%). With four of seven models, the criterion holds at its threshold. It holds in every subset that leaves one model out and in 78.4\% of window-bootstrap resamples. The hidden share reaches 10\% for six of seven models (all except \pithree at 9.3\%), and the criterion holds in 99.3\% of resamples. \Cref{tab:supp-base-t2} gives the other metrics of test-2, which registration v1.2 lists as reported only: $\delta_1$ ranges from 0.601 to 0.659 across the seven counted models, and Depth Pro again answers no ray.

\input{generated/tabS_robust}

\paragraph{Window-level values.}
\Cref{fig:supp-windows} shows both statistics window by window. The hidden share exceeds 10\% in every test-1 window for every model; on test-2 it falls below 10\% in five of the eight windows for \pithree, in four for MoGe-2 (its median over the eight windows, the mean of the two middle ones, is 10.1\%), and in at most one for the other models. The free-space answer rate exceeds 10\% in every test-1 window for VGGT-1B, MoGe-2, DA3, and MASt3R and in none for 3DGS; on test-2 it does so in every window only for MoGe-2 and DA3.

\paragraph{Tolerance.}
\Cref{tab:supp-tau} recomputes both registered statistics with the free-space tolerance $\tau_s$ and the endpoint tolerance $\tauv$ jointly halved and doubled, keeping the stored $b$, $u$, and edge flags fixed. The free-space answer rate decreases as the tolerance grows, since a wider margin shrinks the region counted as free, and the six models above 10\% at the registered tolerance stay above it at both settings (lowest: \pithree at 12.8\% with both tolerances doubled), while 3DGS stays at 6.0--7.9\%. The hidden share changes by at most 0.3 percentage points.

\paragraph{Edge radius, number of inputs, and query grid.}
\Cref{tab:supp-robust} varies the three further settings registered in v1. Changing the edge radius of the \stE label from 2 pixels to 1 or 4 moves the free-space answer rate by at most 1.1 points and the hidden share by at most 1.3 points; the same six models stay above 10\%, and no hidden share falls below 12.7\%. With four input views instead of eight (every second view, so the evidence can only shrink and no ray turns from \stH to \stV), the window-median share of \stV rays among scored rays drops from 46.4\% to 37.1\%, and the hidden share of the four exploration models rises from 13.9--24.4\% to 16.5--31.3\%. Their free-space answers, pooled over the rays of a window (a descriptive variant of the registered per-query statistic), move in both directions (MoGe-2 from 42.4\% to 28.9\%, VGGT-1B from 19.3\% to 23.0\%) and stay above 10\%. A denser query grid of $180{\times}320$ rays (stride 6) was run on two development and two test-1 windows, as registered, so its values (hidden share 12.4--26.4\% and pooled free-space answers 12.1--34.6\% for the six learned models) are not comparable window by window with the test-1 results.

\paragraph{3DGS on hidden endpoints.}
3DGS answers 98\% of the \stH rays of test-1. Its median ratio of predicted to true depth is 2.58 on these rays, 61\% of its answers lie beyond twice the true depth, and on \stV rays the median ratio is 0.987.

%% file: generated/tabS_reference.tex
\begin{table*}[t]
\centering
\caption{\textbf{Reference answers on test-1.} Window medians of four constructed answers and of the training-free evidence wrapper, which answers the predicted upper bound $\hat b$ where it is finite and the model's depth elsewhere. $\delta_1$, FV$_{\mathrm{masked}}$, and BV$_{\mathrm{partial}}$ as in \cref{tab:supp-base}. \emph{\stV{} rays}: fraction answered, and $\delta_1$ among the answered ones. \emph{Scored rays}: fraction without an answer, and fraction with a front violation.}
\label{tab:supp-reference}
\tabsetup
\begin{adjustbox}{max width=\textwidth, gstore width=\tabwidth}
\begin{tabular}{?l|C{34pt}|C{40pt}C{40pt}|C{48pt}C{48pt}|C{42pt}|C{42pt}?}
\tabtop
\tabfirst{2}{Answer} & \multirow{2}{*}{$\delta_1$} & \multicolumn{2}{c|}{\stV{} rays} & \multicolumn{2}{c|}{Scored rays} & \multirow{2}{*}{FV$_{\mathrm{masked}}$} & \multirow{2}{*}{BV$_{\mathrm{partial}}$} \\
\cline{3-6}
 & & answered & $\delta_1$ & unanswered & front viol. & & \\
\tabhead
No answer on any ray & 0.000 & 0.000 & \emptycell & 1.000 & 0.000 & 0.000 & 1.000 \\
Constant depth per query$^\dagger$ & 0.758 & 1.000 & 0.864 & 0.000 & 0.258 & 0.063 & 0.441 \\
Upper bound $b$ where finite & 0.460 & 1.000 & 1.000 & 0.518 & 0.000 & 0.000 & 0.000 \\
True depth & 1.000 & 1.000 & 1.000 & 0.000 & 0.000 & 0.000 & 0.000 \\
\hline
Evidence wrapper, VGGT-1B & 0.563 & 0.991 & 0.983 & 0.420 & 0.137 & 0.040 & 0.153 \\
Evidence wrapper, \pithree & 0.563 & 0.998 & 0.993 & 0.425 & 0.031 & 0.029 & 0.058 \\
\tabbottom
\end{tabular}
\end{adjustbox}
\tabnotes{$^\dagger$The median true depth of the query's \stV{} rays. It reads the ground truth and is a diagnostic of the metric, not a method.}
\end{table*}

%% file: generated/tabS_twins.tex
\begin{table*}[t]
\centering
\caption{\textbf{One prediction, two worlds.} Scores on the divergence region $D$ of 40 certified twins (medians over twins; $D$ has a median of 1{,}160 rays). \emph{Cov.}: fraction of $D$ answered. \emph{Agree}: fraction of rays whose identified-set judgment is the same in both worlds. \emph{Err.}: mean error in the worse of the two worlds on the answered rays of $D$, next to the floor below which no answer can go (half the gap between the two first hits). AbsRel and Err.\ are medians over the 25--30 twins in which a learned model answers rays of $D$ (all 40 for 3DGS). Below the line: Depth Pro, which fails the inclusion check. Gray: confirmation models.}
\label{tab:twins}
\tabsetup
\begin{adjustbox}{max width=\textwidth}
\begin{tabular}{?r|C{36pt}|C{36pt}C{36pt}|C{36pt}C{36pt}|C{36pt}|C{36pt}C{36pt}?}
\tabtop
\tabfirst{2}{Model} & \multirow{2}{*}{Cov.} & \multicolumn{2}{c|}{$\delta_1$} & \multicolumn{2}{c|}{AbsRel} & \multirow{2}{*}{Agree} & \multicolumn{2}{c?}{Err.\ (m)} \\
\cline{3-6}\cline{8-9}
 & & world A & world B & world A & world B & & model & floor \\
\tabhead
VGGT-1B~\cite{wang2025vggt} & 0.37 & 0.286 & 0.000 & 0.036 & 2.05 & 1.000 & 128.9 & 63.3 \\
\pithree~\cite{wang2025pi3} & 0.29 & 0.246 & 0.000 & 0.026 & 2.29 & 1.000 & 131.5 & 63.7 \\
MapAnything~\cite{keetha2025mapanything} & 0.27 & 0.256 & 0.000 & 0.114 & 2.12 & 1.000 & 125.8 & 63.0 \\
MoGe-2~\cite{wang2025moge2} & 0.26 & 0.221 & 0.000 & 0.059 & 2.26 & 1.000 & 124.2 & 64.2 \\
\rowcolor{tabgroup} DA3~\cite{lin2025da3} & 0.37 & 0.332 & 0.000 & 0.068 & 2.03 & 1.000 & 123.3 & 63.4 \\
\rowcolor{tabgroup} MASt3R~\cite{leroy2024mast3r} & 0.32 & 0.253 & 0.000 & 0.107 & 2.18 & 1.000 & 141.3 & 62.9 \\
\rowcolor{tabgroup} 3DGS~\cite{kerbl20233dgs} & 1.00 & 0.286 & 0.000 & 1.403 & 4.21 & 1.000 & 337.0 & 56.5 \\
\hline
Depth Pro~\cite{bochkovskii2025depthpro} & 0.00 & 0.000 & 0.000 & \emptycell & \emptycell & 1.000 & \emptycell & \emptycell \\
\tabbottom
\end{tabular}
\end{adjustbox}
\end{table*}

%% file: generated/tabS_strata.tex
\begin{table*}[t]
\centering
\caption{\textbf{Metrics by query stratum on test-1.} Queries are assigned to the interpolation, extrapolation, and low-evidence strata before any ground truth is read (73, 80, and 55 queries). Each value is the median over the ten windows of the per-window median over the queries of the stratum, with the metrics of \cref{tab:supp-base}. Gray: confirmation models. Best of the seven counted models in red, second underlined (by the unrounded values).}
\label{tab:supp-strata}
\tabsetup
\begin{adjustbox}{max width=\textwidth}
\begin{tabular}{?r|C{27pt}C{27pt}C{27pt}|C{27pt}C{27pt}C{27pt}|C{27pt}C{27pt}C{27pt}|C{27pt}C{27pt}C{27pt}?}
\tabtop
\tabfirst{2}{Model} & \multicolumn{3}{c|}{$\delta_1$ $\uparrow$} & \multicolumn{3}{c|}{FV$_{\mathrm{masked}}$ $\downarrow$} & \multicolumn{3}{c|}{FV$_{\mathrm{partial}}$ $\downarrow$} & \multicolumn{3}{c?}{BV$_{\mathrm{partial}}$ $\downarrow$} \\
\cline{2-13}
 & Interp. & Extrap. & Low ev. & Interp. & Extrap. & Low ev. & Interp. & Extrap. & Low ev. & Interp. & Extrap. & Low ev. \\
\tabhead
VGGT-1B~\cite{wang2025vggt} & 0.785 & 0.542 & 0.272 & \secondval{0.091} & \secondval{0.040} & \secondval{0.011} & \secondval{0.199} & \secondval{0.255} & \secondval{0.310} & 0.154 & 0.168 & 0.088 \\
\pithree~\cite{wang2025pi3} & \secondval{0.791} & 0.561 & 0.269 & \bestval{0.051} & \bestval{0.030} & \bestval{0.011} & \bestval{0.029} & \bestval{0.040} & \bestval{0.057} & \bestval{0.046} & \bestval{0.050} & \bestval{0.075} \\
MapAnything~\cite{keetha2025mapanything} & 0.755 & 0.558 & \bestval{0.321} & 0.125 & 0.048 & 0.015 & 0.428 & 0.379 & 0.582 & 0.422 & 0.433 & 0.232 \\
MoGe-2~\cite{wang2025moge2} & 0.784 & 0.538 & 0.256 & 0.135 & 0.066 & 0.017 & 0.717 & 0.551 & 0.544 & 0.153 & 0.182 & 0.090 \\
\rowcolor{tabgroup} DA3~\cite{lin2025da3} & \bestval{0.831} & \bestval{0.601} & 0.311 & 0.197 & 0.107 & 0.034 & 0.870 & 0.852 & 0.811 & \secondval{0.097} & \secondval{0.090} & \secondval{0.081} \\
\rowcolor{tabgroup} MASt3R~\cite{leroy2024mast3r} & 0.790 & \secondval{0.581} & \secondval{0.320} & 0.164 & 0.095 & 0.026 & 0.603 & 0.662 & 0.630 & 0.165 & 0.196 & 0.205 \\
\rowcolor{tabgroup} 3DGS~\cite{kerbl20233dgs} & 0.772 & 0.536 & 0.267 & 0.135 & 0.075 & 0.026 & 0.466 & 0.554 & 0.522 & 0.138 & 0.148 & 0.182 \\
\tabbottom
\end{tabular}
\end{adjustbox}
\end{table*}

%% file: generated/tabS_base_t2.tex
\begin{table*}[t]
\centering
\caption{\textbf{Metrics on test-2}, the replication set, whose other main-table metrics registration v1.2 lists as reported only. Window medians over the eight windows (upper box) with 95\% window-bootstrap intervals (lower box), defined as in \cref{tab:supp-base}; the confidence AUROC (deviation D9) and the \stH{}-within-$\tauv$ column of that table are computed outside the frozen evaluator and only for test-1. Gray: confirmation models. Columns with an arrow: best of the seven counted models in red, second underlined (by the unrounded values).}
\label{tab:supp-base-t2}
\tabsetup
\setlength{\tabcolsep}{4.6pt}
\begin{adjustbox}{max width=\textwidth, gstore width=\tabwidth}
\begin{tabular}{?r|cc|ccc|c?}
\tabtop
\tabfirst{2}{Model} & \multicolumn{2}{c|}{Standard} & \multicolumn{3}{c|}{Identified-set violations} & \multirow{2}{*}{UCR} \\
\cline{2-6}
\multicolumn{1}{?c|}{} & $\delta_1$ $\uparrow$ & AbsRel $\downarrow$ & FV$_{\mathrm{masked}}$ $\downarrow$ & FV$_{\mathrm{partial}}$ $\downarrow$ & BV$_{\mathrm{partial}}$ $\downarrow$ & \\
\tabhead
VGGT-1B & 0.630 & 0.040 & \secondval{0.011} & \secondval{0.044} & 0.635 & 0.174 \\
\pithree & 0.602 & \bestval{0.024} & \bestval{0.007} & \bestval{0.009} & 0.395 & 0.348 \\
MapAnything & \bestval{0.659} & 0.092 & 0.035 & 0.244 & 0.382 & 0.124 \\
MoGe-2 & 0.618 & \secondval{0.039} & 0.046 & 0.609 & 0.113 & 0.198 \\
\rowcolor{tabgroup} DA3 & \secondval{0.650} & 0.049 & 0.068 & 0.651 & \bestval{0.039} & 0.175 \\
\rowcolor{tabgroup} MASt3R & 0.640 & 0.051 & 0.070 & 0.502 & \secondval{0.096} & 0.118 \\
\rowcolor{tabgroup} 3DGS & 0.601 & 0.511 & 0.057 & 0.489 & 0.205 & 0.264 \\
\hline
Depth Pro$^\dagger$ & 0.000 & \emptycell & 0.000 & 0.000 & 1.000 & 0.000 \\
\tabbottom
\tabgap
\tabtop
\multicolumn{1}{?c|}{95\% interval} & $\delta_1$ & AbsRel & FV$_{\mathrm{masked}}$ & FV$_{\mathrm{partial}}$ & BV$_{\mathrm{partial}}$ & UCR \\
\tabhead
VGGT-1B & \ci{[0.546, 0.677]} & \ci{[0.025, 0.094]} & \ci{[0.006, 0.025]} & \ci{[0.013, 0.095]} & \ci{[0.331, 0.875]} & \ci{[0.102, 0.498]} \\
\pithree & \ci{[0.545, 0.664]} & \ci{[0.012, 0.040]} & \ci{[0.001, 0.018]} & \ci{[0.004, 0.016]} & \ci{[0.016, 0.822]} & \ci{[0.110, 0.577]} \\
MapAnything & \ci{[0.562, 0.708]} & \ci{[0.046, 0.145]} & \ci{[0.016, 0.041]} & \ci{[0.165, 0.589]} & \ci{[0.189, 0.767]} & \ci{[0.076, 0.226]} \\
MoGe-2 & \ci{[0.552, 0.673]} & \ci{[0.032, 0.068]} & \ci{[0.031, 0.064]} & \ci{[0.466, 0.771]} & \ci{[0.057, 0.245]} & \ci{[0.071, 0.258]} \\
\rowcolor{tabgroup} DA3 & \ci{[0.620, 0.707]} & \ci{[0.030, 0.056]} & \ci{[0.056, 0.133]} & \ci{[0.604, 0.894]} & \ci{[0.020, 0.073]} & \ci{[0.073, 0.341]} \\
\rowcolor{tabgroup} MASt3R & \ci{[0.543, 0.701]} & \ci{[0.036, 0.103]} & \ci{[0.016, 0.111]} & \ci{[0.200, 0.912]} & \ci{[0.034, 0.189]} & \ci{[0.044, 0.437]} \\
\rowcolor{tabgroup} 3DGS & \ci{[0.556, 0.652]} & \ci{[0.473, 0.920]} & \ci{[0.052, 0.080]} & \ci{[0.440, 0.529]} & \ci{[0.155, 0.303]} & \ci{[0.228, 0.334]} \\
\tabbottom
\end{tabular}
\end{adjustbox}
\tabnotes{$^\dagger$Fails the inclusion check and answers no query ray, as on test-1.}
\end{table*}

%% file: generated/tabS_robust.tex
\begin{table*}[t]
\centering
\caption{\textbf{Robustness to evaluation settings} (values in \%; blue: the registered setting). Left: the two registered statistics on test-1 with the edge radius of the \stE{} label at 1, 2, and 4 pixels. Right: $K{=}4$ inputs (every second input view, identified sets recomputed; exploration models, ten test-1 windows) and a $180{\times}320$ query grid (two development and two test-1 windows, so these columns are not paired with the others; 3DGS not run). These two runs report free-space answers pooled over the rays of a window, a descriptive variant of the registered per-query statistic; the $K{=}8$ column gives the pooled value of the registered setting ($90{\times}160$ rays) for comparison. N/A: not run.}
\label{tab:supp-robust}
\tabsetup
\begin{adjustbox}{max width=\textwidth}
\begin{tabular}{?r|C{24pt}C{24pt}C{24pt}|C{24pt}C{24pt}C{24pt}|C{34pt}C{34pt}C{34pt}|C{34pt}C{34pt}?}
\tabtop
\tabfirst{3}{Model} & \multicolumn{6}{c|}{Edge radius $r_e$ (pixels)} & \multicolumn{5}{c?}{Inputs and query grid} \\
\cline{2-12}
 & \multicolumn{3}{c|}{Free-space answers} & \multicolumn{3}{c|}{Hidden share} & \multicolumn{3}{c|}{Free-space answers, pooled} & \multicolumn{2}{c?}{Hidden share} \\
\cline{2-12}
 & 1 & \cellcolor{tabours}2 & 4 & 1 & \cellcolor{tabours}2 & 4 & \cellcolor{tabours}$K{=}8$ & $K{=}4$ & $180{\times}320$ & $K{=}4$ & $180{\times}320$ \\
\tabhead
VGGT-1B~\cite{wang2025vggt} & 19.4 & \cellcolor{tabours}19.9 & 19.2 & 16.2 & \cellcolor{tabours}15.6 & 14.7 & \cellcolor{tabours}19.3 & 23.0 & 15.8 & 18.9 & 13.6 \\
\pithree~\cite{wang2025pi3} & 14.7 & \cellcolor{tabours}15.0 & 16.1 & 14.6 & \cellcolor{tabours}13.9 & 12.7 & \cellcolor{tabours}15.5 & 11.6 & 12.1 & 16.5 & 12.4 \\
MapAnything~\cite{keetha2025mapanything} & 14.7 & \cellcolor{tabours}14.5 & 13.8 & 25.0 & \cellcolor{tabours}24.4 & 23.6 & \cellcolor{tabours}15.2 & 14.6 & 23.1 & 31.3 & 26.4 \\
MoGe-2~\cite{wang2025moge2} & 38.6 & \cellcolor{tabours}39.6 & 40.6 & 15.6 & \cellcolor{tabours}14.9 & 13.8 & \cellcolor{tabours}42.4 & 28.9 & 34.0 & 16.8 & 12.5 \\
DA3~\cite{lin2025da3} & 33.5 & \cellcolor{tabours}33.1 & 32.4 & 23.1 & \cellcolor{tabours}22.4 & 21.4 & \cellcolor{tabours}32.8 & N/A & 32.2 & N/A & 19.1 \\
MASt3R~\cite{leroy2024mast3r} & 27.7 & \cellcolor{tabours}27.3 & 26.6 & 23.4 & \cellcolor{tabours}22.8 & 21.8 & \cellcolor{tabours}27.1 & N/A & 34.6 & N/A & 19.7 \\
3DGS~\cite{kerbl20233dgs} & 7.7 & \cellcolor{tabours}7.3 & 6.8 & 20.2 & \cellcolor{tabours}19.4 & 18.1 & \cellcolor{tabours}7.2 & N/A & N/A & N/A & N/A \\
\tabbottom
\end{tabular}
\end{adjustbox}
\end{table*}

%% file: supp/D_head.tex
\section{\texorpdfstring{\head}{ID-Head}: details and secondary results}
\label{supp:head}

\paragraph{Features.}
\Cref{tab:supp-features} defines the 18 per-ray features. The predicted evidence applies the construction of \cref{sec:evidence} to the backbone's own input-view depths with the frozen settings of the training-free evidence wrapper (no confidence threshold, tolerance scaled by 2). Distances are divided by the flight altitude, non-finite values are set to 0, and explicit flags keep the information that a value was missing.

\begin{table}[tp]
\centering
\caption{\textbf{Per-ray input features of \head.} Hats mark quantities computed from the backbone's predictions. The predicted evidence forms the ten evidence channels that the variant of \cref{tab:variants} leaves out.}
\label{tab:supp-features}
\tabsetup
\setlength{\tabcolsep}{4pt}
\begin{adjustbox}{max width=\columnwidth}
\begin{tabular}{?l|l?}
\tabtop
\multicolumn{1}{?c|}{Feature} & \multicolumn{1}{c?}{Definition} \\
\tabhead
\multicolumn{2}{?l?}{\textit{Backbone readout (6 channels)}} \\
range & readout depth along the ray \\
covered & the readout is finite \\
confidence & backbone confidence of the splatted point \\
confidence present & a confidence value exists \\
second gap & gap between the first two z-buffer layers \\
point count & $\log(1 + \text{points splatted into the pixel})$ \\
\hline
\multicolumn{2}{?l?}{\textit{Predicted evidence (10 channels)}} \\
$\hat a$ & end of the free prefix of the ray in $\hat\free$ \\
$\hat b$ & first predicted observed surface along the ray \\
$\hat b$ finite & the ray meets a predicted observed surface \\
$\hat u / \hat L$ & predicted evidence gap \\
answer in $\hat\free$ & the readout lies in predicted free space \\
category (5) & one-hot evidence category under $\hat\free$, $\hat b$ \\
\hline
\multicolumn{2}{?l?}{\textit{Geometry (2 channels)}} \\
ray direction & downward component of the ray \\
nearest camera & distance to the nearest input camera \\
\tabbottom
\end{tabular}
\end{adjustbox}
\end{table}

\begin{table*}[tp]
\centering
\caption{\textbf{Validation gate G1$'$.} The 20 validation windows, seeds 20261025--20261027. Violations count answered \stH{} rays that violate the evidence at \stV{} recall 0.807. Green: the row meets the condition of the column, pink: it does not. The gate requires two of three seeds to meet all three conditions.}
\label{tab:supp-gate}
\tabsetup
\begin{adjustbox}{max width=\textwidth}
\begin{tabular}{?l|l|C{50pt}C{102pt}|C{60pt}C{48pt}C{76pt}|C{40pt}?}
\tabtop
\multicolumn{2}{?c|}{\multirow{2}{*}{Version}} & \multicolumn{2}{c|}{(a$'$) ranking} & \multicolumn{3}{c|}{(b$'$) violations} & (c$'$) \\
\cline{3-8}
\multicolumn{2}{?c|}{} & AUROC & Diff.\ (CI lower) & Count & Per 1{,}000 \stH & Ratio \ci{[CI]} & $\delta_1$ \\
\hline
\multicolumn{2}{?l|}{Threshold} & $\ge 0.8451$ & $>0$ & $\le$ 44{,}153 & $\le$ 19.8 & upper $<1$ & $\ge 0.9596$ \\
\tabhead
v1a & three seeds & \cellmet 0.856--0.857 & \cellmet +0.051--0.052 (0.043--0.045) & \cellnotmet 58{,}052--58{,}415 & \cellnotmet 26.1--26.2 & \cellmet 0.92 \ci{[0.90, 0.94]} & \cellmet 0.974 \\
\hline
\multirow{3}{*}{v1b} & seed 20261025 & \cellmet 0.898 & \cellmet +0.098 (0.076) & \cellnotmet 45{,}707 & \cellnotmet 20.5 & \cellmet 0.725 \ci{[0.635, 0.810]} & \cellmet 0.976 \\
 & seed 20261026 & \cellmet 0.899 & \cellmet +0.098 (0.076) & \cellnotmet 45{,}302 & \cellnotmet 20.3 & \cellmet 0.718 \ci{[0.640, 0.796]} & \cellmet 0.977 \\
 & seed 20261027 & \cellmet 0.903 & \cellmet +0.097 (0.078) & \cellmet 43{,}057 & \cellmet 19.3 & \cellmet 0.683 \ci{[0.595, 0.768]} & \cellmet 0.977 \\
\tabbottom
\end{tabular}
\end{adjustbox}
\end{table*}

\paragraph{Architecture and training.}
The U-Net applies two $3{\times}3$ convolutions with 16 channels, three stride-2 convolutions to 32, 64, and 128 channels, each followed by a $3{\times}3$ convolution, and three transposed convolutions back to full resolution, each concatenated with the encoder features of the same scale. A $1{\times}1$ convolution produces the logit. All activations are ReLU, and the network has 349{,}073 parameters; each output logit has a receptive field of about $47{\times}47$ rays. Training uses the full maps of the 1{,}562 training queries: \stV rays have target 1, and every other ray has target 0, including the \stE and \stX rays that evaluation excludes. Each query's loss is weighted by its stratum (interpolation 0.814, extrapolation 0.938, low evidence 4.0), and \stH rays whose backbone answer violates the evidence receive weight 3. These violations are computed once on the training queries and serve only as weights. One step processes one full $90{\times}160$ query map. We use Adam with learning rate $10^{-3}$, clip the gradient norm at 5, train for 10 epochs, and keep the checkpoint with the lowest validation loss. The answer scale $s = 1.0096$ is the median ratio of true to readout depth over the answered \stV rays of the training windows. The lowest validation loss is 0.370--0.374 for the five registered seeds, 0.376--0.378 without the evidence channels, 0.383--0.385 with the evidence channels only, and 0.379--0.380 and 0.394--0.395 with 50 and 25 of the 100 training windows.

\input{generated/tabS_ablation}

\paragraph{Development and the validation gate.}
Three head versions were trained and evaluated on validation windows only: v0, a per-ray MLP that also predicted depth and an interval; v1a, a per-ray MLP with a single logit; and v1b, the U-Net reported in the main text. v0 did not pass the validation gate G1 defined for it. A revised gate G1$'$ was defined after the validation results of v0 were known and before v1a and v1b were trained: (a$'$)~validation AUROC of at least 0.8451 (0.05 above the point-count signal's validation AUROC of 0.7951), with the CI lower bound of the difference above 0; (b$'$)~at \stV recall 0.807 (the validation \stV recall of the evidence wrapper), at most 0.7 times the violations of the best training-free gate, which was the evidence wrapper (63{,}076 violations), with the CI upper bound of the ratio below 1; (c$'$)~$\delta_1$ of at least 0.9596 on the answered \stV rays (0.01 below the backbone's validation $\delta_1$ of 0.9696). The gate is passed if at least two of three seeds meet all three conditions. Six earlier runs of v1a and v1b that deviated from the gate's specification (weight 3 on every \stH ray instead of the violating ones, no stratum weights, and a U-Net with a single downsampling) were excluded as gate candidates before the gate was evaluated. \Cref{tab:supp-gate} shows the outcome: every seed of both versions meets (a$'$) and (c$'$), one v1b seed meets (b$'$), and the gate is not passed. Under the interpretation fixed together with G1$'$, this outcome still allows the head to be the main-text method, with its AUROC gains and violation changes reported as measured. v1b, with two additional seeds trained to the same specification, was then registered in v1.2 with test hypotheses that differ from the gate: each test set uses its own best training-free comparator, violations are averaged over the recall range, and each hypothesis is decided separately.

\paragraph{Validation ablations.}
\Cref{tab:supp-ablation} presents the validation results behind \cref{sec:analysis}. More training windows raise AUROC and lower the ratio of violating answers to the evidence wrapper. The variant without evidence channels reaches the ratio of the registered head at a slightly lower AUROC, and the variant with the evidence channels only answers more violating rays. Cyclically shifting the evidence channels along the rays of each query breaks their spatial alignment with the other eight channels and lowers AUROC to 0.651--0.670. The drop shows that the trained head uses the evidence channels at inference time. Because retraining without them recovers most of the AUROC (0.889--0.895), the test does not measure their unique contribution.

\input{generated/tabS_operating}

\input{generated/tab5_variants}

\input{generated/tabS_pi3}

\input{generated/tabS_mechanism}

\paragraph{Results known at registration.}
Registration v1.2 lists the validation and development results that were known when it was written. On validation, the five v1b seeds reach AUROC 0.898--0.903 and answer 10.4--11.1 violating rays per 1{,}000 \stH rays averaged over recall, against 18.5 for the evidence-gap gate. On the two development windows at 150\,m, v1b reaches AUROC 0.849--0.857 on one and 0.782--0.802 on the other, against 0.746 and 0.752 for the point count, and answers 10.8--11.9 violating rays per 1{,}000 \stH rays averaged over recall, against 13.7 for the point count, the best gate there. Below \stV recall 0.6, the confidence gate answers fewer violating rays than v1b on these windows. After the protocol had been fixed and before it was tagged, a dry run of it on the development windows met the ranking criterion in five of five seeds and the violation criterion in two of five (ratios 0.79--0.87 against the point count), and it showed the variant without evidence channels ahead of v1b (7.0--7.5 vs.\ 10.8--11.9 violating answers per 1{,}000 \stH rays averaged over recall). The registered head was kept unchanged, and the variant remained a secondary result.

\begin{figure*}[tp]
\centering
\includegraphics[width=\textwidth]{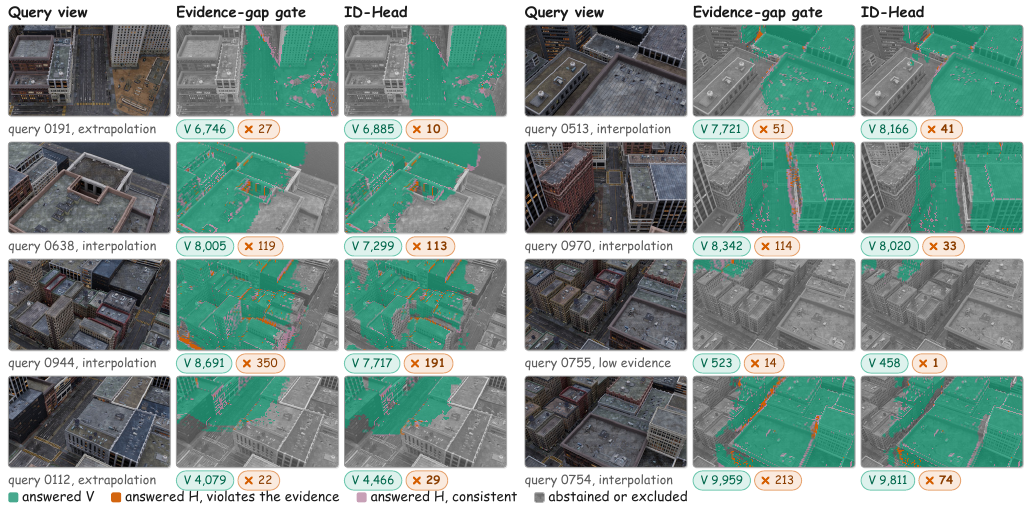}
\caption{\textbf{Where \head abstains: eight further test-1 queries} at \stV recall 0.807, as \cref{fig:qual}, one from each test-1 window that the main paper does not show. Untinted rays are abstentions or excluded \stE and \stX rays. Badges: answered \stV rays and violating answers.}
\label{fig:supp-abstain}
\end{figure*}

\begin{figure*}[tp]
\centering
\includegraphics[width=\textwidth]{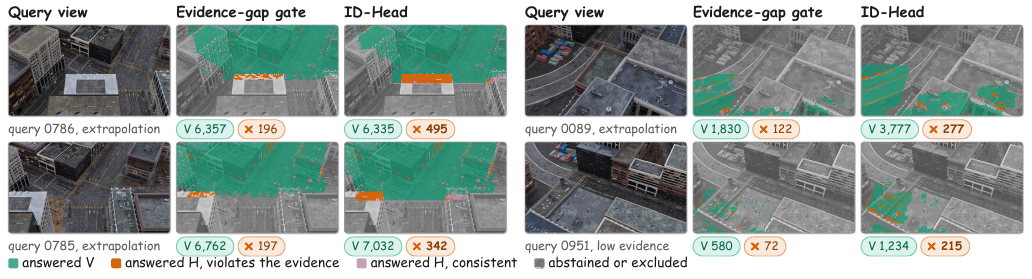}
\caption{\textbf{Failure cases of \head}: the four test-1 queries with the largest excess of violating answers over the evidence-gap gate, at \stV recall 0.807, drawn as \cref{fig:qual}. Untinted rays are abstentions or excluded \stE and \stX rays. Badges: answered \stV rays and violating answers.}
\label{fig:supp-failures}
\end{figure*}

\begin{figure}[tp]
\centering
\includegraphics[width=\columnwidth]{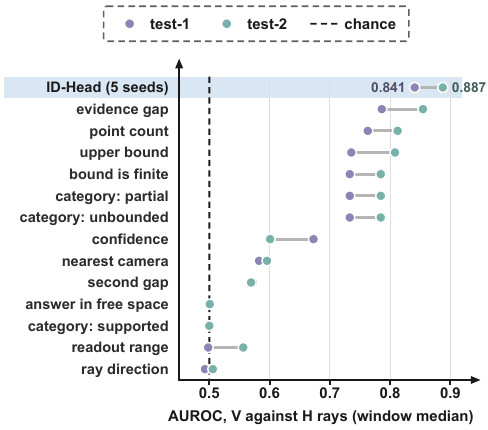}
\caption{\textbf{Single signals.} Window-median AUROC of \stV against \stH rays for each single signal on VGGT-1B, sorted by test-1, with \head (mean of five seeds) in the first row. The evidence signals are computed from the predicted evidence. Three rows (bound is finite, category: partial, category: unbounded) have identical values, and the supported category is at chance on both test sets.}
\label{fig:supp-features}
\end{figure}

\begin{figure}[tp]
\centering
\includegraphics[width=\columnwidth]{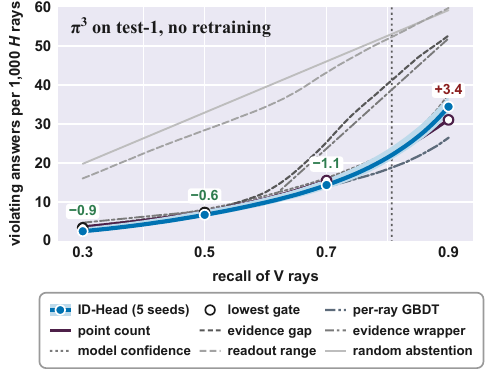}
\caption{\textbf{Violation--recall curves on \pithree} (test-1, no retraining), drawn as \cref{fig:curves}. The band spans the five seeds of \head. Labels: mean of the five seeds of \head minus the strongest (lowest-violation) training-free gate at that recall (legend: lowest gate; green: fewer violating answers, red: more). Dotted: recall 0.807.}
\label{fig:supp-pi3}
\end{figure}

\begin{figure*}[tp]
\centering
\includegraphics[width=\textwidth]{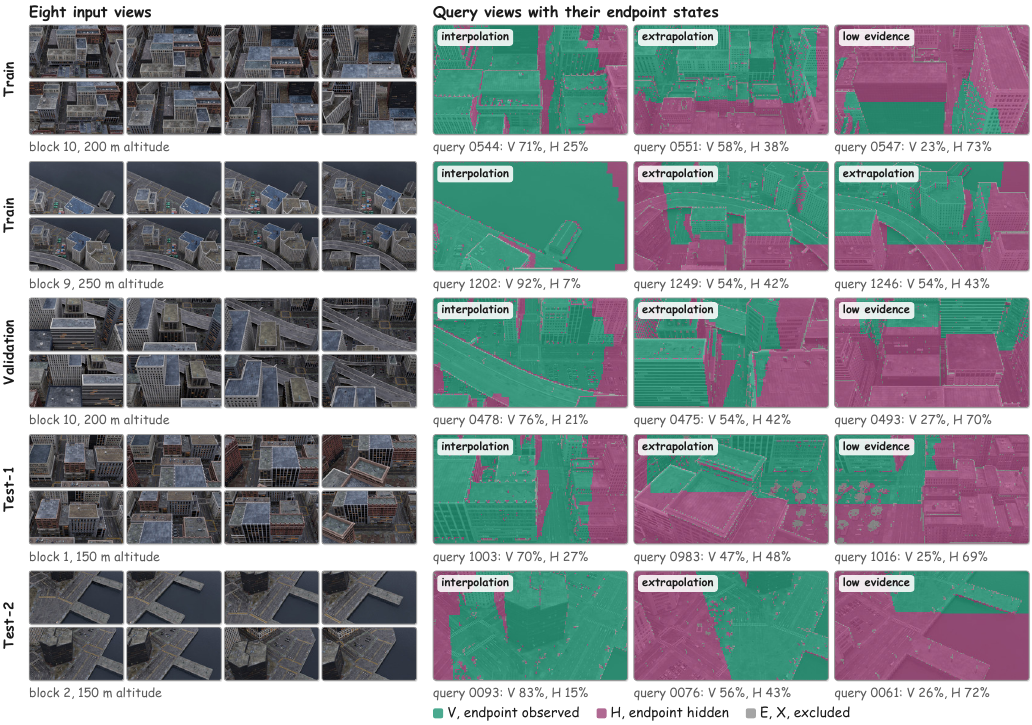}
\caption{\textbf{One window of each split and block}: training windows of blocks 10 and 9, a validation window, and a window of each test set. Left: the eight input views. Right: the endpoint states of one query per stratum (the training window of block~9 has no low-evidence query, so it shows a second extrapolation query), with the shares of \stV and \stH rays.}
\label{fig:supp-splits}
\end{figure*}

\paragraph{Operating points.}
\Cref{tab:supp-operating} reports the violating answers of every gate and head at the three operating points that registration v1.2 lists as secondary results. With thresholds fixed on validation at \stV recall 0.807, the recall of \head on each test set follows that of the evidence wrapper: 0.61--0.63 against 0.631 on test-1, and 0.86--0.87 against 0.862 on test-2. At \stV recall 0.807 on test-1, \head answers 11{,}789--12{,}234 violating \stH rays without a predicted upper bound against 24{,}955 for the evidence-gap gate, and 18{,}677--19{,}546 against 18{,}736 on rays with a finite $\hat b$. On test-2 the corresponding counts are 231--416 against 1{,}218 and 6{,}386--7{,}177 against 8{,}192. The evidence-gap gate is the comparator here and in \cref{fig:qual} because it is the best gate of both test sets averaged over recall; at P2 alone, the point count answers fewer violating rays on test-1. The registered evaluation uses fractional tie-breaking at the threshold to match the target \stV recall exactly, which is also how a binary gate such as the evidence wrapper reaches a fixed recall. Because the evidence-gap score is quantized, each tied ray on test-1 is counted with weight 0.185 at P2. The counts above, like \cref{fig:qual}, instead treat all tied rays as abstentions, which puts the gate at \stV recall 0.806 on test-1 and 0.803 on test-2 (0.812 and 0.807 if all tied rays are answered); accordingly, they sum to 43{,}691 and 9{,}410 rather than the registered 43{,}840 and 9{,}539 in \cref{tab:supp-operating}.

\paragraph{Typical queries.}
At \stV recall 0.807 and with seed 20261025, \head incurs fewer violating answers than the evidence-gap gate on 152 of the 208 test-1 queries, more on 55, and ties on 1 (median difference, gate minus \head: 30); on test-2 it incurs fewer on 80 of the 123 queries, more on 31, and ties on 12 (median difference: 10). \Cref{fig:supp-abstain} repeats \cref{fig:qual} on eight further test-1 queries with the same checkpoint and operating point. A fixed rule picks one query from each test-1 window that the main paper does not show: the query at the median difference of the window, among its queries not used for other examples. Ordered by the difference, each of the eight lies within two places of the median of all queries of its window. Of the eight, \head incurs more violating answers than the gate only on query~0112 (29 against 22).

\paragraph{Failure cases.}
\Cref{fig:supp-failures} shows the four test-1 queries with the largest excess of violating answers over the evidence-gap gate. On the extrapolation queries 0786 and 0785, \head answers about as many \stV rays as the gate and 1.7--2.5 times as many violating ones. On the extrapolation query 0089 and the low-evidence query 0951, it answers about twice as many \stV rays and 2.3--3.0 times as many violating ones (badges in \cref{fig:supp-failures}); on these two queries the shared operating point lets it answer more of the query overall.

\paragraph{Single signals.}
\Cref{fig:supp-features} ranks the 13 single signals among which the ranking comparator is chosen (\cref{sec:protocol}) by their window-median AUROC. The predicted evidence gap is the strongest on both test sets (0.786 and 0.854), followed by the splatted point count (0.763 and 0.812) and the predicted upper bound (0.736 and 0.808). The backbone confidence reaches 0.673 on test-1 and 0.601 on test-2 (with rays pooled within each window, as for every signal here; the benchmark evaluator in \cref{tab:supp-base} takes the median over the queries of a window and yields 0.692 on test-1). Both readout range and ray direction remain within 0.06 AUROC of chance (0.50). \head reaches 0.841 and 0.887. Per window, the mean AUROC of the five seeds of \head ranges from 0.762 to 0.873 on test-1 and from 0.764 to 0.940 on test-2, and the seeds differ by at most 0.018.

\paragraph{Variants and the ensemble.}
\Cref{tab:variants} compares the registered head with the variant trained without evidence channels and with the per-ray GBDT on all three settings, as discussed in \cref{sec:analysis}. Behind its seed counts for ranking, the window-median AUROC on test-1 is 0.843--0.853 for the variant without evidence channels (difference to the evidence gap $+0.0702$ to $+0.0716$, CI lower bounds $-0.0048$ to $+0.0128$) and 0.843 for the GBDT ($+0.0489$, lower bound $+0.0148$), against 0.831--0.845 for the registered head ($+0.0501$ to $+0.0547$). The GBDT reaches 0.890 on test-2 (difference $+0.0156$, lower bound $+0.0041$) and 0.836 on \pithree (difference to the point count $+0.0394$, lower bound $+0.0121$). The mean-logit ensemble of the five seeds, a registered secondary result, meets both criteria on both test sets: window-median AUROC 0.846 and 0.893, AUROC difference $+0.0571$ [$+0.0263$, $+0.0971$] and $+0.0345$ [$+0.0091$, $+0.0745$], and violation ratio 0.747 [0.579, 0.878] and 0.613 [0.501, 0.724] (95\% paired window-bootstrap intervals).

\paragraph{Transfer to \pithree.}
\Cref{tab:supp-pi3} reports the transfer to \pithree without retraining seed by seed: two of the five seeds meet the ranking criterion and none meets the violation criterion. The mean-logit ensemble meets the ranking criterion ($+0.0382$ [$+0.0012$, $+0.0605$], under the frozen bootstrap stream of \cref{supp:prereg}) and not the violation criterion (0.894 [0.746, 1.043]), with a window-median AUROC of 0.833. \Cref{fig:supp-pi3} draws the violation--recall curves of the five seeds. At \stV recall 0.3, 0.5, and 0.7, their mean lies 0.6--1.1 violating rays per 1{,}000 \stH rays below the strongest (lowest-violation) training-free gate, and at each of these recalls one seed lies above it; at recall 0.9 \head answers 34.45 against 31.05 for the point count, and the per-ray GBDT answers 26.44.

\paragraph{Mechanism split.}
\Cref{tab:supp-mechanism} splits the AUROC and the violating answers of the five seeds by the predicted bound and by query stratum, a secondary result of registration v1.2. On test-1, the AUROC is lower on rays with a finite $\hat b$ (0.730--0.739) than on rays without one (0.767--0.775); on test-2 the two are close (0.794--0.800 and 0.799--0.807). On both test sets and at every operating point, most violating answers fall on rays with a finite $\hat b$. Pooled in the same way, the AUROC on \pithree is 0.790--0.825, 0.770--0.812, and 0.759--0.795 on interpolation, extrapolation, and low-evidence queries.

%% file: generated/tabS_ablation.tex
\begin{table*}[t]
\centering
\caption{\textbf{Validation ablations of \head.} The 20 validation windows (368 queries), ranges over seeds. Violating answers count answered \stH{} rays that violate the evidence at \stV{} recall 0.807, and the ratio divides them by the 63{,}076 of the evidence wrapper, as in \cref{tab:supp-gate}; $\delta_1$ is measured on the answered \stV{} rays. Rows 1--2 train the registered head on a subset of the 100 training windows. \emph{Evidence channels only} keeps the ten evidence channels and the coverage flag. The last row evaluates three of the five registered checkpoints after a cyclic shift of the ten evidence channels along the rays of each query, which keeps their values and breaks their alignment with the rays. The counts come from the released evaluation, which scores in single precision, and differ from \cref{tab:supp-gate} by at most one ray. Blue: the registered head.}
\label{tab:supp-ablation}
\tabsetup
\begin{adjustbox}{max width=\textwidth}
\begin{tabular}{?l|C{32pt}C{24pt}|C{48pt}|C{62pt}C{46pt}C{56pt}|C{48pt}?}
\tabtop
\tabfirst{2}{Variant} & \multicolumn{2}{c|}{Training} & \multirow{2}{*}{AUROC} & \multicolumn{3}{c|}{Violating answers} & \multirow{2}{*}{$\delta_1$} \\
\cline{2-3}\cline{5-7}
 & Windows & Seeds & & Count & Per 1{,}000 \stH & Ratio to wrapper & \\
\tabhead
25 training windows & 25 & 3 & 0.885--0.887 & 51{,}544--52{,}933 & 23.1--23.8 & 0.817--0.839 & 0.974 \\
50 training windows & 50 & 3 & 0.893--0.895 & 46{,}938--47{,}469 & 21.1--21.3 & 0.744--0.753 & 0.975--0.976 \\
\hline
\rowcolor{tabours} \head (registered) & 100 & 5 & 0.898--0.903 & 43{,}056--45{,}707 & 19.3--20.5 & 0.683--0.725 & 0.976--0.977 \\
\hline
Without evidence channels & 100 & 3 & 0.889--0.895 & 42{,}898--45{,}284 & 19.3--20.3 & 0.680--0.718 & 0.976--0.978 \\
Evidence channels only & 100 & 3 & 0.888--0.893 & 48{,}385--50{,}471 & 21.7--22.7 & 0.767--0.800 & 0.974--0.975 \\
\hline
\head, evidence channels shifted & 100 & 3 & 0.651--0.670 & 109{,}345--112{,}658 & 49.1--50.6 & 1.734--1.786 & 0.970--0.971 \\
\tabbottom
\end{tabular}
\end{adjustbox}
\end{table*}

%% file: generated/tabS_operating.tex
\begin{table*}[t]
\centering
\caption{\textbf{Operating points.} Answered \stH{} rays that violate the evidence at three operating points (counts over all windows of a test set). P1: the \stV{} recall of the evidence wrapper (0.631 on test-1, 0.862 on test-2). P2: \stV{} recall 0.807. P3: the threshold of each gate or head fixed on validation, with the \stV{} recall it reaches on the test set. Ranges are over seeds. At P1 and P2, the lowest count is in red and the second lowest underlined (means over seeds); P3 counts are taken at different recalls and are not ranked.}
\label{tab:supp-operating}
\tabsetup
\begin{adjustbox}{max width=\textwidth}
\begin{tabular}{?l|C{64pt}C{64pt}|C{64pt}C{64pt}?}
\tabtop
\tabfirst{2}{Test-1} & \multicolumn{2}{c|}{Fixed \stV{} recall} & \multicolumn{2}{c?}{Fixed threshold} \\
\cline{2-5}
 & P1 $\downarrow$ & P2 $\downarrow$ & P3 recall & P3 \\
\tabhead
No gate (random abstention) & 47{,}694 & 61{,}004 & 0.813 & 61{,}488 \\
Model confidence & 33{,}771 & 52{,}438 & 0.916 & 71{,}005 \\
Readout range & 43{,}200 & 57{,}983 & 0.907 & 66{,}716 \\
Splatted point count & 24{,}308 & 37{,}856 & 0.758 & 33{,}730 \\
Evidence wrapper ($\hat b<\infty$) & 20{,}808 & 47{,}402 & 0.631 & 20{,}808 \\
Evidence gap $\hat u/\hat L$ & 19{,}719 & 43{,}840 & 0.681 & 25{,}889 \\
\hline
\rowcolor{tabours} \head (5 seeds) & 15{,}826--16{,}538 & 30{,}781--31{,}335 & 0.612--0.633 & 14{,}705--16{,}705 \\
Mean-logit ensemble & \secondval{15{,}902} & \secondval{30{,}625} & 0.619 & 15{,}231 \\
Without evidence channels (3 seeds) & \bestval{14{,}132--15{,}502} & \bestval{27{,}238--28{,}827} & 0.741--0.749 & 21{,}567--23{,}608 \\
Per-ray GBDT & 18{,}953 & 33{,}466 & 0.616 & 18{,}076 \\
\tabbottom
\tabgap
\tabtop
\tabfirst{2}{Test-2} & \multicolumn{2}{c|}{Fixed \stV{} recall} & \multicolumn{2}{c?}{Fixed threshold} \\
\cline{2-5}
 & P1 $\downarrow$ & P2 $\downarrow$ & P3 recall & P3 \\
\tabhead
No gate (random abstention) & 25{,}776 & 24{,}123 & 0.812 & 24{,}279 \\
Model confidence & 29{,}296 & 27{,}765 & 0.817 & 28{,}661 \\
Readout range & 23{,}668 & 21{,}257 & 0.929 & 26{,}913 \\
Splatted point count & 17{,}821 & 14{,}919 & 0.763 & 13{,}192 \\
Evidence wrapper ($\hat b<\infty$) & 13{,}692 & 12{,}814 & 0.862 & 13{,}692 \\
Evidence gap $\hat u/\hat L$ & 13{,}348 & 9{,}539 & 0.823 & 10{,}288 \\
\hline
\rowcolor{tabours} \head (5 seeds) & \secondval{9{,}908--10{,}528} & \secondval{6{,}716--7{,}593} & 0.862--0.868 & 10{,}048--10{,}962 \\
Mean-logit ensemble & \bestval{9{,}773} & \bestval{6{,}703} & 0.865 & 9{,}942 \\
Without evidence channels (3 seeds) & 10{,}898--11{,}067 & 8{,}257--8{,}347 & 0.833--0.841 & 9{,}443--9{,}798 \\
Per-ray GBDT & 11{,}620 & 9{,}337 & 0.822 & 9{,}842 \\
\tabbottom
\end{tabular}
\end{adjustbox}
\end{table*}

%% file: generated/tab5_variants.tex
\begin{table*}[t]
\centering
\caption{\textbf{What the head relies on.} Violation ratio to the best training-free gate of each setting (range over seeds) and the number of seeds that meet the ranking and the violation criteria of \cref{tab:method}. All heads are trained on VGGT-1B features; the \pithree columns apply them without retraining. Only the registered head was part of the registered hypotheses; the other rows are secondary results. The variant without evidence channels was registered as a secondary result on test-1 only; its test-2 and \pithree columns come from the second implementation of \cref{supp:prereg}, not from the registered run. Lowest ratio in red, second underlined.}
\label{tab:variants}
\tabsetup
\begin{adjustbox}{max width=\textwidth}
\begin{tabular}{?l|c|ccc|ccc|ccc?}
\tabtop
\tabfirst{2}{Selector} & \multirow{2}{*}{Seeds} & \multicolumn{3}{c|}{Test-1} & \multicolumn{3}{c|}{Test-2} & \multicolumn{3}{c?}{\pithree, test-1 (no retraining)} \\
\cline{3-11}
 & & Ratio $\downarrow$ & Ranking & Violations & Ratio $\downarrow$ & Ranking & Violations & Ratio $\downarrow$ & Ranking & Violations \\
\tabhead
\rowcolor{tabours} \head (registered) & 5 & \secondval{0.75--0.78} & 5/5 & 5/5 & \bestval{0.63--0.71} & 5/5 & 5/5 & 0.90--1.07 & 2/5 & 0/5 \\
\quad without evidence channels & 3 & \bestval{0.67--0.72} & 2/3 & 3/3 & \secondval{0.81--0.86} & 0/3 & 0/3 & \bestval{0.72--0.79} & 3/3 & 3/3 \\
Per-ray GBDT, same features & 1 & 0.87 & 1/1 & 0/1 & 0.99 & 1/1 & 0/1 & \secondval{0.86} & 1/1 & 1/1 \\
\tabbottom
\end{tabular}
\end{adjustbox}
\end{table*}

%% file: generated/tabS_pi3.tex
\begin{table*}[t]
\centering
\caption{\textbf{Zero-shot transfer of \head to \pithree.} Test-1, without retraining (answer scale 0.9880 from the training windows). The comparators are \pithree's own training-free signals, and its splatted point count is the comparator of both criteria. Difference: median over windows of per-window AUROC differences. Violations: violating \stH{} answers per 1{,}000 \stH{} rays, averaged over \stV{} recall 0.30--0.90. Green: the seed meets the criterion of \cref{tab:method}, pink: it does not.}
\label{tab:supp-pi3}
\tabsetup
\begin{adjustbox}{max width=\textwidth}
\begin{tabular}{?l|C{52pt}C{52pt}C{52pt}C{52pt}C{52pt}?}
\tabtop
\tabfirst{2}{Ranking: AUROC of \stV{} vs.\ \stH{} $\uparrow$} & \multicolumn{5}{c?}{Seed} \\
\cline{2-6}
 & 20261025 & 20261026 & 20261027 & 20261028 & 20261029 \\
\tabhead
Splatted point count & \multicolumn{5}{c?}{0.807} \\
\hline
\rowcolor{tabours} \head & 0.797 & 0.839 & 0.834 & 0.824 & 0.822 \\
\rowcolor{tabours} \quad difference & +0.0092 & +0.0403 & +0.0408 & +0.0324 & +0.0246 \\
\rowcolor{tabours} \quad 95\% CI lower bound & \cellnotmet $-$0.0290 & \cellmet +0.0071 & \cellmet +0.0002 & \cellnotmet $-$0.0114 & \cellnotmet $-$0.0029 \\
\rowcolor{tabours} \quad 95\% CI upper bound & +0.0315 & +0.0627 & +0.0578 & +0.0522 & +0.0474 \\
\tabbottom
\tabgap
\tabtop
\tabfirst{2}{Selective answering: violations $\downarrow$} & \multicolumn{5}{c?}{Seed} \\
\cline{2-6}
 & 20261025 & 20261026 & 20261027 & 20261028 & 20261029 \\
\tabhead
No gate (random abstention) & \multicolumn{5}{c?}{39.50} \\
Model confidence & \multicolumn{5}{c?}{13.96} \\
Readout range & \multicolumn{5}{c?}{36.36} \\
Splatted point count & \multicolumn{5}{c?}{12.93} \\
Evidence wrapper ($\hat b<\infty$) & \multicolumn{5}{c?}{19.00} \\
Evidence gap $\hat u/\hat L$ & \multicolumn{5}{c?}{19.77} \\
\hline
\rowcolor{tabours} \head & 13.81 & 11.58 & 11.60 & 12.04 & 12.53 \\
\rowcolor{tabours} \quad ratio to the best gate & 1.069 & 0.896 & 0.898 & 0.932 & 0.969 \\
\rowcolor{tabours} \quad 95\% CI lower bound & 0.868 & 0.761 & 0.763 & 0.783 & 0.799 \\
\rowcolor{tabours} \quad 95\% CI upper bound & \cellnotmet 1.292 & \cellnotmet 1.022 & \cellnotmet 1.034 & \cellnotmet 1.092 & \cellnotmet 1.140 \\
\tabbottom
\end{tabular}
\end{adjustbox}
\end{table*}

%% file: generated/tabS_mechanism.tex
\begin{table*}[t]
\centering
\caption{\textbf{Mechanism split of \head}, a secondary result of registration v1.2: the rays split by whether the predicted upper bound $\hat b$ is finite and by query stratum. AUROC of \stV{} against \stH{} rays, pooled over the answered rays of each test set; violating answers: answered \stH{} rays that violate the evidence at the operating points P1--P3 of \cref{tab:supp-operating} (counts over all windows). Ranges over the five seeds.}
\label{tab:supp-mechanism}
\tabsetup
\begin{adjustbox}{max width=\textwidth}
\begin{tabular}{?l|cccc|cccc?}
\tabtop
\tabfirst{2}{Rays} & \multicolumn{4}{c|}{Test-1} & \multicolumn{4}{c?}{Test-2} \\
\cline{2-9}
 & AUROC & P1 & P2 & P3 & AUROC & P1 & P2 & P3 \\
\tabhead
$\hat b$ finite & 0.730--0.739 & 13{,}112--13{,}880 & 18{,}677--19{,}546 & 12{,}548--13{,}975 & 0.794--0.800 & 8{,}977--9{,}484 & 6{,}386--7{,}177 & 9{,}083--9{,}784 \\
$\hat b = +\infty$ & 0.767--0.775 & 2{,}523--3{,}134 & 11{,}789--12{,}234 & 2{,}157--3{,}170 & 0.799--0.807 & 710--1{,}044 & 231--416 & 732--1{,}178 \\
\hline
Interpolation & 0.839--0.845 & 5{,}714--6{,}091 & 11{,}639--12{,}065 & 5{,}535--6{,}141 & 0.893--0.895 & 2{,}648--2{,}947 & 2{,}078--2{,}313 & 2{,}641--3{,}040 \\
Extrapolation & 0.815--0.822 & 9{,}143--9{,}643 & 16{,}514--16{,}815 & 8{,}522--9{,}727 & 0.872--0.876 & 5{,}674--6{,}045 & 3{,}619--4{,}372 & 5{,}758--6{,}282 \\
Low evidence & 0.816--0.823 & 652--1{,}080 & 2{,}557--2{,}742 & 565--1{,}095 & 0.845--0.855 & 1{,}321--1{,}570 & 782--908 & 1{,}344--1{,}640 \\
\tabbottom
\end{tabular}
\end{adjustbox}
\end{table*}

%% file: supp/E_data.tex
\section{Data and release}
\label{supp:data}

\paragraph{Source data.}
All scenes come from the aerial captures of MatrixCity~\cite{li2023matrixcity}: blocks 1 and 2 at 150\,m for the test sets, and blocks 9 and 10 at 200--250\,m for training and validation. Camera positions and depth are converted from the dataset's units to meters. After conversion, neighboring frames reproject onto each other with a median relative depth error of 0.014\% for the adjacent aerial pair of the first convention check; in the later check recorded in \texttt{IDSet3D/data/conventions.json}, the median over two pairs is 0.0067\%. Frames are keyed by block and frame index, because frame indices repeat across blocks. \Cref{tab:benchmark} summarizes the size and endpoint-state composition of each split, and Fig.~2d of the main paper shows the per-query distribution of \stH-ray shares. \Cref{fig:supp-splits} shows one window of each split and block, with its eight inputs and three of its queries.

\input{generated/tabS_statistics}

\paragraph{Release.}
The \href{https://huggingface.co/datasets/wolfkingttf/IDSet3D/tree/1ff8c0d1985e31cbaa6e8863ba3ff8003bd61a73}{archived IDSet-3D dataset revision} provides the per-ray evidence and endpoint labels, the \head features, model predictions, and the frozen result files. The \href{https://huggingface.co/wolfkingttf/IDSet3D-IDHead/tree/570a01bd03eb289f2fe2a9778ad5f19af10f4ed2}{archived ID-Head model revision} contains the five registered checkpoints, ablation and control models, and validation thresholds and training logs. Each repository includes a SHA-256 manifest. Both releases use CC BY-NC 4.0. The \href{https://huggingface.co/datasets/BoDai/MatrixCity}{original MatrixCity archive} provides the input images and depth maps.

\paragraph{Reported estimates.}
The data repository includes the \href{https://huggingface.co/datasets/wolfkingttf/IDSet3D/tree/1ff8c0d1985e31cbaa6e8863ba3ff8003bd61a73/eval/v1.0/full}{test-1 benchmark outputs}, the \href{https://huggingface.co/datasets/wolfkingttf/IDSet3D/tree/1ff8c0d1985e31cbaa6e8863ba3ff8003bd61a73/eval/c83_formal/test2/scores}{test-2 benchmark scores}, and the \href{https://huggingface.co/datasets/wolfkingttf/IDSet3D/resolve/1ff8c0d1985e31cbaa6e8863ba3ff8003bd61a73/eval/c83_formal/test1/method_VGGT-1B.json}{test-1} and \href{https://huggingface.co/datasets/wolfkingttf/IDSet3D/resolve/1ff8c0d1985e31cbaa6e8863ba3ff8003bd61a73/eval/c83_formal/test2/method_VGGT-1B.json}{test-2} ID-Head results in JSON. These files record the window summaries, comparator scores, and seed-level intervals underlying \cref{tab:findings,tab:method}.

%% file: generated/tabS_statistics.tex
\begin{table}[t]
\centering
\caption{\textbf{\bench statistics.} Each query view contributes 14{,}400 rays. \stV{} and \stH{} are the observed- and hidden-endpoint rays; the excluded rays are edge-ambiguous (\stE) or evidence-inconsistent (\stX). No ray misses the city (\stN). The two test blocks are used by no training or validation window. Endpoint states in \%.}
\label{tab:benchmark}
\tabsetup
\setlength{\tabcolsep}{2.5pt}
\begin{adjustbox}{max width=\columnwidth}
\begin{tabular}{?l|c|c|ccc|ccc?}
\tabtop
\tabfirst{2}{Split} & \multirow{2}{*}{Block} & \multirow{2}{*}{Alt.\ (m)} & \multicolumn{3}{c|}{Size} & \multicolumn{3}{c?}{Endpoint states} \\
\cline{4-9}
 & & & Win. & Queries & Rays & \stV & \stH & Excl. \\
\tabhead
Train & 9, 10 & 200--250 & 100 & 1{,}562 & 22.5M & 63.3 & 33.5 & 3.2 \\
Val & 10 & 200 & 20 & 368 & 5.3M & 55.2 & 42.0 & 2.8 \\
Test-1 & 1 & 150 & 10 & 208 & 3.0M & 46.3 & 50.1 & 3.5 \\
Test-2 & 2 & 150 & 8 & 123 & 1.8M & 53.2 & 44.1 & 2.7 \\
\hline
Total & 1, 2, 9, 10 & 150--250 & 138 & 2{,}261 & 32.6M & 59.9 & 37.0 & 3.1 \\
\tabbottom
\end{tabular}
\end{adjustbox}
\end{table}